\documentclass[10pt,twocolumn,letterpaper]{article}

\usepackage[pagenumbers]{wacv}
\usepackage{bm}
\usepackage[accsupp]{axessibility}
\definecolor{wacvblue}{rgb}{0.21,0.49,0.74}
\usepackage[pagebackref,breaklinks,colorlinks,allcolors=wacvblue]{hyperref}

\usepackage{etoolbox}
\makeatletter
\patchcmd{\NAT@citexnum}
{\let\NAT@last@num\NAT@num}
{%
  \MakeLinkTarget[cite]{}%
  \Hy@backout{\@citeb\@extra@b@citeb}%
  \let\NAT@last@num\NAT@num
}
{}
{\PackageWarning{wacv-backref}{Failed to patch natbib back references}}
\makeatother

\def\wacvPaperID{1518}
\def\confName{WACV}
\def\confYear{2027}

\title{Structure from Collision via Hypothesis-Space Reasoning}

\author{
  Takuhiro Kaneko\\
  NTT, Inc.
}

\begin{document}

\twocolumn[{
  \renewcommand\twocolumn[1][]{#1}
  \maketitle
  \vspace{-9mm}
  \begin{center}
    \includegraphics[width=\textwidth]{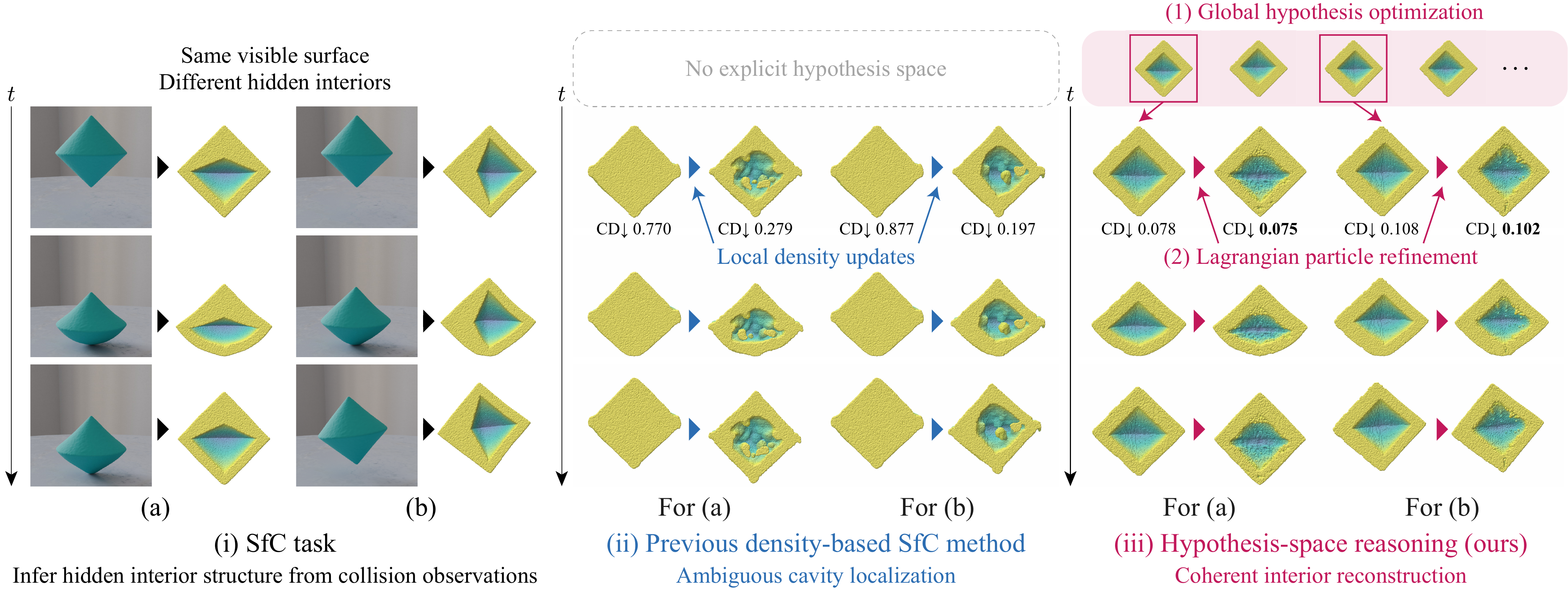}
  \end{center}
  \vspace{-5mm}
  \captionof{figure}{Overview of Structure from Collision (SfC) via hypothesis-space reasoning.
    (i) SfC infers the hidden interior structure from collision observations of objects with the same visible surface but different interiors.
    (ii) A previous density-based SfC method~\cite{TKanekoCVPR2025} uses local density updates without an explicit hypothesis space, leading to ambiguous cavity localization.
    (iii) Our method optimizes explicit structural hypotheses globally and refines the selected hypothesis as a Lagrangian particle system, yielding coherent interior reconstruction.}
  \label{fig:teaser}
  \vspace{5mm}
}]

\addtocontents{toc}{\protect\setcounter{tocdepth}{-1}}

\begin{abstract}
  Knowledge of the hidden interior structure of an object, including the presence and location of cavities, is essential for predictive simulation, robotic manipulation, and nondestructive inspection. However, such cavities leave no direct signature on the static visible surface, making collision-induced deformation one of the few cues used to infer their presence. Structure from Collision (SfC) addresses this problem by recovering the hidden interior structure of an object from multiview observations of a collision event. SfC is challenging for two reasons: hidden cavities are only indirectly observable through surface motion, and voids correspond to the absence of material, making direct optimization using standard density representations difficult. The density-based SfC baseline inherits both difficulties. Surface-matching gradients do not directly indicate the cavity location, while density gradients can be weak or uninformative inside or near void regions. Therefore, we reformulate SfC as hypothesis-space reasoning, representing hidden cavities as explicit structural hypotheses whose variables encode available structural assumptions rather than relying on implicit density updates. A global stage performs gradient-free optimization over this hypothesis space using forward-only material point method (MPM) simulation. A local stage then refines the selected hypothesis as a Lagrangian particle system through gradient-based optimization using differentiable MPM simulation. On the SfC benchmark, our method improves hidden-structure reconstruction and future prediction over the density-based SfC baseline. Interior-topology stress tests and robustness analyses characterize its scope and limitations.%
  \footnote{The project page is available at \url{https://www.kecl.ntt.co.jp/people/kaneko.takuhiro/projects/sfc-hypothesis/}.}
\end{abstract}

\vspace{-2mm}
\section{Introduction}
\label{sec:introduction}

Vision-based 3D reconstruction is important for applications that require physical understanding, such as predictive simulation, robotic manipulation, and nondestructive inspection.
Recent methods have recovered detailed visible geometry and appearance from images or videos~\cite{BMildenhallECCV2020,BKerblTOG2023}.
However, visible geometry alone is insufficient to explain the physical behavior governed by internal mass distribution.
For example, as shown in Fig.~\ref{fig:teaser}(i), two objects with identical external surfaces but different hidden cavities are indistinguishable from static visual observations, yet can respond differently to the same collision.

Structure from Collision (SfC)~\cite{TKanekoCVPR2025} addresses this limitation by using collision-induced motion to infer the hidden interior structure from multiview observations.
Because post-collision surface motion reflects the internal mass distribution, cavities indistinguishable from the static appearance can be inferred from the deformation.
However, this inverse problem is difficult for two related reasons: hidden cavities are only indirectly observable, and void regions are difficult to optimize directly with standard density representations because they represent the absence of material.
As shown in Fig.~\ref{fig:challenge}, the post-collision surface-motion discrepancy between a hidden-cavity object and its filled initialization can suggest an internal correction at a location different from the true cavity.
A previous density-based SfC method~\cite{TKanekoCVPR2025} inherits both difficulties. Without an explicit hypothesis space, surface-matching gradients may suggest corrections away from the true cavity, while density updates inside or near void regions can be weak or uninformative, leading to ambiguous cavity localization (Fig.~\ref{fig:teaser}(ii)).

To address these limitations, we reformulate SfC as hypothesis-space reasoning, replacing local density updates with explicit structural hypotheses evaluated against the observed collision response.
This representation directly encodes available structural assumptions in the hypothesis space rather than indirectly through a density field.
We instantiate this formulation in two complementary stages (Fig.~\ref{fig:teaser}(iii)).
The global stage performs gradient-free optimization over the hypothesis space using forward-only material point method (MPM) simulation~\cite{DSulskyCMAME1994,YHuTOG2018}, while the local stage refines the selected hypothesis as a Lagrangian particle system through gradient-based optimization with differentiable MPM simulation~\cite{YHuICLR2020}, recovering details beyond the initial hypothesis parameterization.
This two-stage design separates coarse structural reasoning from fine geometric refinement.

On the SfC benchmark, the proposed method improves hidden-structure reconstruction and future prediction over the density-based SfC baseline~\cite{TKanekoCVPR2025}.
Interior-topology stress tests and robustness analyses further characterize its scope and limitations.

Our contributions are summarized as follows:
\begin{itemize}
\item We identify a structural failure mode of density-based SfC. Indirect surface cues and void regions make local density gradients unreliable for cavity localization.
\item We reformulate SfC as hypothesis-space reasoning, representing hidden cavities as explicit structural hypotheses rather than relying on implicit density updates.
\item We propose a two-stage framework that combines gradient-free global hypothesis optimization with gradient-based Lagrangian particle refinement, separating hypothesis-level structural selection from particle-level geometric refinement.
\item We demonstrate improved reconstruction and future prediction over the density-based SfC baseline~\cite{TKanekoCVPR2025}, with interior-topology stress tests and robustness analyses clarifying the scope and limitations of the method.
\end{itemize}

\section{Related Work}
\label{sec:related_work}

\noindent\textbf{3D reconstruction.}
Classical structure-from-motion~\cite{CTomasiIJCV1992,NSnavelyTOG2006} and multiview stereo~\cite{YFurukawaTPAMI2009,SSeitzCVPR2006} recover surface geometry from images.
Neural 3D representations, such as neural radiance fields (NeRF)~\cite{BMildenhallECCV2020} and 3D Gaussian splatting~\cite{BKerblTOG2023}, further improve the reconstruction of visible geometry and appearance.
However, these methods primarily model visible properties and do not explicitly target the hidden interior structures that govern the physical response.

\smallskip
\noindent\textbf{Dynamic 3D reconstruction.}
3D reconstruction methods have been extended to dynamic and nonrigid scenes~\cite{ZLiCVPR2021,APumarolaCVPR2021,GGafniCVPR2021,ETretschkICCV2021,KParkICCV2021,ZYangICLR2024,ZYangCVPR2024,JLuiten3DV2024,ZLuCVPR2024,GWuCVPR2024,KKatsumataECCV2024,JYanCVPR2025}.
These methods typically learn deformation fields, canonical representations, and time-varying Gaussian primitives for temporally coherent reconstruction and dynamic view synthesis.
However, they primarily model visible motion and appearance, and do not explicitly target hidden interior structures revealed only indirectly through physical interaction.

\smallskip
\noindent\textbf{Physics-informed modeling and system identification.}
Recent studies have incorporated physical constraints or differentiable simulators to enforce plausible dynamics or estimate physical quantities from visual or multimodal observations~\cite{YFengCVPR2024,MChuTOG2022,SGuanICML2022,XLiICLR2023,TKanekoCVPR2024,JAbouWACV2024,TXieCVPR2024,YJiangTOG2024,RZQiuECCV2024,FLiuArXiv2024,TZhangECCV2024,JNiNeurIPS2024,THuangAAAI2025,JAbouCoRL2024,LZhongECCV2024,JCaiNeurIPS2024,ZLiuCVPR2025,YZhaoICCV2025,HJiangICCV2025,FVasileNeurIPS2025,CLiuICLR2026,FShaikhCVPR2026,NWadaCVPR2026}.
However, these methods typically assume a fixed interior, even though internal mass distribution can strongly affect physical behavior.

\smallskip
\noindent\textbf{Structure from Collision.}
SfC~\cite{TKanekoCVPR2025} aims to recover internal cavity structures from collision-induced deformation.
The previous method, SfC-NeRF~\cite{TKanekoCVPR2025}, optimizes volumetric densities in an Eulerian representation such that the simulated surface dynamics match the observed motion.
Although effective in some settings, this density-based formulation relies on local density updates without an explicit structural hypothesis space, which can result in ambiguous cavity localization.
In contrast, by representing hidden cavities as explicit structural hypotheses, we first optimize them globally and then refine the selected hypothesis as a Lagrangian particle system.

\section{Problem Setup and Failure Analysis}
\label{sec:problem}

\subsection{Problem Definition}
\label{subsec:problem_definition}

We consider the problem of inferring the hidden interior structure from visual observations of a collision event.
Given calibrated multiview videos $\{\mathcal{I}_i(t_j) \mid i=1,\ldots,M,\; j=1,\ldots,N\}$, where $M$ is the number of views and $N$ is the number of time steps, we obtain two observation streams: a time-varying visible surface $\mathcal{S}(t_j)$ and image-space object masks $\mathcal{A}_i(t_j)$.
The surface $\mathcal{S}(t_j)$ can be reconstructed from multiview observations using an off-the-shelf dynamic 3D reconstruction method~\cite{GWuCVPR2024,ZYangCVPR2024,JCaiNeurIPS2024}.
The masks $\mathcal{A}_i(t_j)$ can be extracted using video matting~\cite{SLinCVPR2021} or video object segmentation methods~\cite{NRaviICLR2025}.
These observations constrain the external geometry and motion of an object but do not directly reveal hidden cavities.
Thus, SfC aims to recover the hidden interior structure that shapes the collision response of an object rather than merely explaining the visible surface evolution.

We represent the object as a particle-based continuum.
Let $\mathcal{P}(t_1)$ denote the ground-truth initial particle configuration and $\hat{\mathcal{P}}(t_1)$ denote its estimate.
Given $\hat{\mathcal{P}}(t_1)$ and physical material parameters $\Theta_{\mathrm{phys}}$, a physics simulator (e.g., MPM~\cite{DSulskyCMAME1994,YHuTOG2018}) predicts the evolving particle configurations $\hat{\mathcal{P}}(t_j)$.
From these simulated configurations, we extract predicted surfaces $\hat{\mathcal{S}}(t_j)$ and render predicted masks $\hat{\mathcal{A}}_i(t_j)$.
Thus, SfC can be formulated as an inverse problem: estimating an initial particle configuration $\hat{\mathcal{P}}(t_1)$ that captures the hidden interior structure whose simulated surface and mask evolution match the observed collision dynamics.

Unless otherwise stated, we treat the physical material parameters $\Theta_{\mathrm{phys}}$, such as the Young's modulus, Poisson's ratio, and mass density, as given.
This controlled setting isolates the structural ambiguity from material-parameter ambiguity.
However, specifying these physical parameters does not specify the cavity; its location, shape, and topology remain unknown.
To assess the sensitivity of this assumption, we evaluate representative robustness settings in Section~\ref{subsec:robustness}, with additional relaxed assumptions and physical-parameter error analyses in Appendices~\ref{subsec:relaxing_assumptions} and~\ref{subsec:physical_parameter_errors}.

Under this formulation, a natural approach is to optimize the internal representation so that the predicted surfaces and masks match the observations.
However, density-based local optimization is structurally mismatched to hidden-cavity inference, as analyzed in the following subsection.

\begin{figure}[t]
  \centering
  \includegraphics[width=0.99\linewidth]{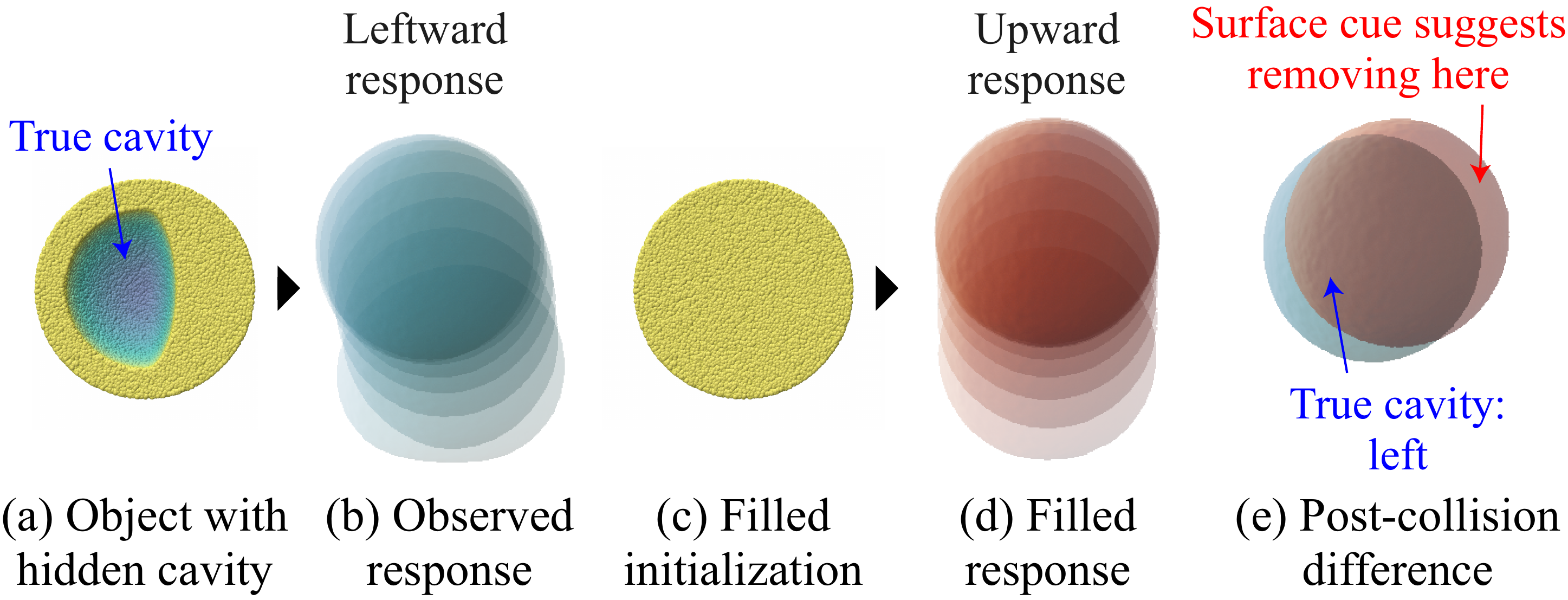}
  \vspace{-1mm}
  \caption{Misleading surface-motion cue for hidden-cavity localization.
    An object with a left-side cavity (a) produces a leftward post-collision response (b), whereas the filled initialization used by a previous density-based SfC method (c) moves upward (d).
    The resulting post-collision difference (e) suggests the removal of material from the right side, even though the true cavity lies on the left.
    Thus, surface-matching cues alone do not reliably localize hidden cavities and can mislead local density-based optimization.}
  \label{fig:challenge}
  \vspace{-1mm}
\end{figure}

\subsection{Why Density-Based Local Optimization Fails}
\label{subsec:failure_analysis}

Fig.~\ref{fig:challenge} illustrates a representative failure mode.
A hidden cavity can induce a post-collision response whose surface-motion discrepancy points away from the true cavity location.
Thus, matching visible motion does not localize the area from which the material should be removed.

This example reflects two coupled difficulties.
First, hidden cavities are only indirectly observable; surface or silhouette motion does not directly indicate which internal region is responsible.
Second, cavities are void regions (i.e., the absence of material), making it difficult to optimize them directly with density-based representations.
At the gradient level, local updates are driven by surface- or mask-level discrepancies. Near the visible surface, gradients tend to adjust the external motion or silhouettes, while inside or near void regions, little material exists to support informative density updates.
Consequently, local updates can reduce observation losses in directions misaligned with the true cavity, leading to ambiguous reconstructions and misleading minima.

SfC-NeRF~\cite{TKanekoCVPR2025} partially mitigates these issues by optimizing volumetric densities and applying volume annealing to escape the local minima.
Nevertheless, its optimization remains tied to local density updates starting from the filled initialization and lacks explicit space for structural alternatives.
Therefore, it can remain sensitive to initialization and may converge to cavity configurations that reduce local observation errors while remaining structurally inconsistent.
Together, these limitations motivate our formulation: we represent candidate cavities as explicit structural hypotheses and evaluate them before local particle-level refinement.

\begin{figure*}[t]
  \centering
  \includegraphics[width=0.99\textwidth]{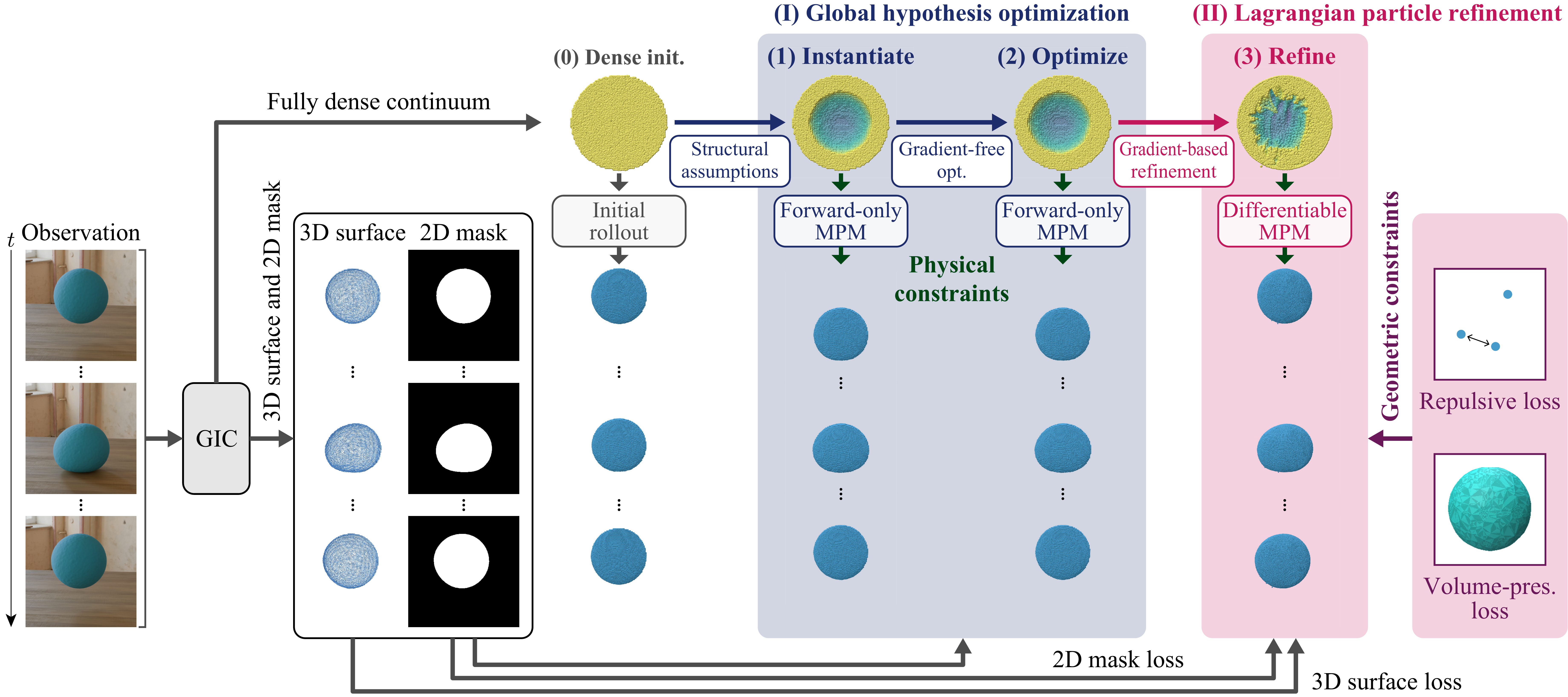}
  \vspace{-1mm}
  \caption{Overview of hypothesis-space reasoning for SfC.
    (0) Dense particle initialization is reconstructed from multiview observations following GIC~\cite{JCaiNeurIPS2024}.
    The initial rollout shows the dynamics before hypothesis optimization.
    (I) Global hypothesis optimization consists of (1) instantiating cavity hypotheses and (2) scoring them with forward-only MPM simulation using a collision-adjacent loss.
    (II) Lagrangian particle refinement: (3) the selected hypothesis is refined using differentiable MPM simulation with physical and geometric constraints.
    This two-stage framework separates coarse structural reasoning from fine geometric refinement.}
  \label{fig:overview}
  \vspace{-1mm}
\end{figure*}

\section{Method: Hypothesis-Space Reasoning}
\label{sec:method}

\subsection{Overview}
\label{subsec:method_overview}

We instantiate SfC as hypothesis-space reasoning by representing hidden cavities as explicit structural hypotheses rather than relying on implicit density updates.
This shifts the optimization from local density updates to hypothesis-level evaluation in a low-dimensional space that can explicitly encode available structural assumptions.

Our method begins with a dense initialization, followed by two optimization stages (Fig.~\ref{fig:overview}).
Following Gaussian-Informed Continuum (GIC)~\cite{JCaiNeurIPS2024}, we first reconstruct a fully dense particle representation from multiview observations (Fig.~\ref{fig:overview}(0)).
Unlike GIC, which uses this representation for physical-parameter identification under a fixed dense interior, we use it as an initialization for hidden-cavity reasoning.
Stage~I performs gradient-free global hypothesis optimization by instantiating and evaluating cavity hypotheses with forward-only MPM simulation~\cite{DSulskyCMAME1994,YHuTOG2018} (Fig.~\ref{fig:overview}(1)--(2)).
Stage~II performs gradient-based Lagrangian particle refinement of the selected hypothesis with differentiable MPM simulation~\cite{YHuICLR2020} (Fig.~\ref{fig:overview}(3)).

\smallskip
\noindent\textbf{Observation losses.}
For a set of time steps $\mathcal{T}$, we define
\begin{flalign}
  \label{eq:loss_3d}
  \mathcal{L}_{\mathrm{3D}}(\mathcal{T})
  &=
    \frac{1}{|\mathcal{T}|}
    \sum_{t_j \in \mathcal{T}}
    d_{\mathrm{CD}}\!\left(
    \mathcal{S}(t_j),
    \hat{\mathcal{S}}(t_j)
    \right),
  \\
  \label{eq:loss_2d}
  \mathcal{L}_{\mathrm{2D}}(\mathcal{T})
  &=
    \frac{1}{M|\mathcal{T}|}
    \sum_{i=1}^{M}
    \sum_{t_j \in \mathcal{T}}
    \left\|
    \mathcal{A}_i(t_j)
    -
    \hat{\mathcal{A}}_i(t_j)
    \right\|_1.
\end{flalign}
where $d_{\mathrm{CD}}$ denotes the Chamfer distance (CD), and the predicted surfaces and masks are obtained by simulating and rendering the current particle configuration.

\subsection{Structural Hypothesis Space}
\label{subsec:hypothesis_space}

We define a structural hypothesis space $\mathcal{H}$ for candidate hidden cavities.
A structural hypothesis denotes a candidate cavity configuration evaluated and selected during optimization.
Starting from the dense particle initialization $\hat{\mathcal{P}}_{\mathrm{dense}}(t_1)$ (Fig.~\ref{fig:overview}(0)), a hypothesis $h \in \mathcal{H}$ specifies an internal void to instantiate (Fig.~\ref{fig:overview}(1)).
We parameterize $h$ using compact structural variables, such as a cavity shape representation $\tau$ (e.g., an exterior-derived shape or a template index), center position $\bm{c}$, size $s$, and orientation $\bm{R}$: $h = (\tau, \bm{c}, s, \bm{R}) \in \mathcal{H}$.
Given $h$, we remove the specified material region from $\hat{\mathcal{P}}_{\mathrm{dense}}(t_1)$, yielding the candidate particle configuration $\hat{\mathcal{P}}_h(t_1)$.
Unlike per-particle or per-voxel density representations, $\mathcal{H}$ directly parameterizes structural hypothesis variables for the hidden-cavity inference.%
\footnote{This instantiation can be extended with additional geometric variables or physical parameters, as evaluated in Section~\ref{subsec:robustness}.}

Structural assumptions define the hypothesis space by fixing or activating different variables.
We consider three settings with different amounts of structural information.

\smallskip
\noindent\emph{(i) Known size.}
When the cavity size is specified, the search focuses on the cavity position.
The cavity shape is initialized by offsetting inward from the observed external surfaces rather than selected from predefined topology templates.

\smallskip
\noindent\emph{(ii) A set of candidate templates and known center.}
When the cavity center and a set of candidate templates are available, the center is fixed, while the template index and remaining configuration variables, such as orientation, are searched.

\smallskip
\noindent\emph{(iii) Known template and center.}
When both the cavity template and center are specified, the hypothesis space is restricted to the remaining configuration variables, such as orientation.

These settings yield compact, low-dimensional hypothesis spaces for global exploration prior to particle-level refinement.
Their dataset-specific instantiations are described in Section~\ref{subsec:experimental_setup} and Appendix~\ref{subsec:model_details}.

\subsection{Global Hypothesis Optimization}
\label{subsec:global_stage}

Given the structural hypothesis space $\mathcal{H}$, the global stage selects a cavity hypothesis by evaluating candidate structures with forward-only simulation (Fig.~\ref{fig:overview}(2)).
Because each hypothesis removes material from the dense particle configuration, it induces a discrete structural change that is not well suited to local particle-gradient updates.
Therefore, we use gradient-free hypothesis-level optimization, with coarse-to-fine enumeration as the default for stability and reproducibility.

In other words, for each evaluated hypothesis $h$, we instantiate $\hat{\mathcal{P}}_h(t_1)$ and perform forward-only MPM simulation~\cite{DSulskyCMAME1994,YHuTOG2018}.
We define the base observation score using the 2D mask loss restricted to collision-adjacent frames.
\begin{flalign}
  \label{eq:global_obs_loss}
  \mathcal{L}_{\mathrm{global}}(h)
  =
  \mathcal{L}_{\mathrm{2D}}^{h}(\mathcal{T}_{\mathrm{col}}),
\end{flalign}
where $\mathcal{T}_{\mathrm{col}}$ denotes the frames immediately before and after the collision, and superscript $h$ indicates masks rendered from the simulation initialized by hypothesis $h$.
Restricting the observation loss to 2D masks and collision-adjacent frames keeps the global stage focused on structural plausibility and avoids penalizing long-horizon discrepancies that an unrefined hypothesis cannot resolve.
For the position search, we augment this base score with a displacement regularizer normalized by the object extent to discourage disproportionately large cavity shifts, as detailed in Appendix~\ref{subsubsec:global_hypothesis_optimization}.
We select the lowest-loss hypothesis $h^\ast$ and pass the corresponding particle configuration $\hat{\mathcal{P}}_{h^\ast}(t_1)$ to the local-refinement stage.

\subsection{Lagrangian Particle Refinement}
\label{subsec:local_stage}

Given the selected configuration $\hat{\mathcal{P}}_{h^\ast}(t_1)$, the local stage refines particle positions to recover geometric details beyond the hypothesis parameterization (Fig.~\ref{fig:overview}(3)).
Unlike the global stage, which evaluates discrete cavity hypotheses, this stage performs continuous particle-level optimization while preserving the selected cavity structure.
Following Lagrangian particle optimization~\cite{TKanekoCVPR2024}, we optimize the initial particle positions through differentiable MPM simulation~\cite{YHuICLR2020}, which propagates observation-loss gradients through simulated dynamics.

The local stage uses the full observed sequence $\mathcal{T}_{\mathrm{all}}=\{t_1,\ldots,t_N\}$:
\begin{flalign}
  \label{eq:local_total_loss}
  \mathcal{L}_{\mathrm{local}}
  ={}&
       \mathcal{L}_{\mathrm{3D}}(\mathcal{T}_{\mathrm{all}})
       +
       \lambda_{\mathrm{2D}}\mathcal{L}_{\mathrm{2D}}(\mathcal{T}_{\mathrm{all}})
       \nonumber\\
     &+
       \lambda_{\mathrm{rep}}\mathcal{L}_{\mathrm{rep}}
       +
       \lambda_{\mathrm{vol}}\mathcal{L}_{\mathrm{vol}}.
\end{flalign}
This full-sequence objective complements the collision-adjacent mask loss in the global stage by refining the particle-level geometry and dynamics over the entire trajectory.

To stabilize the refinement around the selected hypothesis, we use two geometry-aware regularization losses.
The repulsive loss is
\begin{flalign}
  \mathcal{L}_{\mathrm{rep}}
  =
  \frac{1}{|\mathcal{E}_{\mathrm{rep}}|}
  \sum_{(i,j)\in\mathcal{E}_{\mathrm{rep}}}
  \left(
  \frac{\|\hat{p}_i(t_1)-\hat{p}_j(t_1)\|}{d_{\mathrm{grid}}}
  -1
  \right)^2,
\end{flalign}
where $\hat{p}_i(t_1)$ denotes the position of the $i$th particle in $\hat{\mathcal{P}}(t_1)$, $\mathcal{E}_{\mathrm{rep}}$ contains nearby particle pairs, and $d_{\mathrm{grid}}$ is the grid spacing.
This loss discourages particle collapse and excessive clustering.
The volume-preservation loss is
\begin{flalign}
  \mathcal{L}_{\mathrm{vol}}
  =
  \left(
  \frac{\mathcal{V}(\hat{\mathcal{P}}(t_1))+\epsilon}
  {\mathcal{V}(\hat{\mathcal{P}}_{\mathrm{ref}}(t_1))+\epsilon}
  -1
  \right)^2,
\end{flalign}
where $\mathcal{V}(\cdot)$ denotes the estimated material volume of the particle configuration, $\hat{\mathcal{P}}_{\mathrm{ref}}(t_1)$ denotes the reference configuration, and $\epsilon$ denotes a small constant.
The reference can either be fixed to the selected hypothesis or updated during refinement.
We use the latter adaptive variant by default.
Together, these losses allow local adjustment while preserving the coarse structure selected in the global stage.
Loss-design ablations are provided in Section~\ref{subsec:ablation}.

\subsection{Implementation and Complexity}
\label{subsec:implementation}

\textbf{Implementation.}
We kept the hypothesis-grid resolution, MPM time step, refinement steps, and loss weights $\lambda_{\mathrm{2D}}$, $\lambda_{\mathrm{rep}}$, and $\lambda_{\mathrm{vol}}$ fixed across experiments unless otherwise stated.
We used the physical parameters $\Theta_{\mathrm{phys}}$ from the previous SfC setting~\cite{TKanekoCVPR2025}, such that the evaluation focuses on structural inference rather than material-parameter estimation.
In relaxed settings, physical parameters can be included as additional variables, as evaluated in Section~\ref{subsec:robustness}.
Additional implementation details are provided in Appendix~\ref{subsec:model_details}.

\smallskip\noindent\textbf{Complexity.}
The global and local stages have different computational profiles.
The global stage evaluates the cavity hypotheses with forward-only MPM simulation~\cite{DSulskyCMAME1994,YHuTOG2018} on collision-adjacent frames without backpropagation.
For an exhaustive grid with $n$ samples per dimension in a $d$-dimensional hypothesis space, the cost scales as $O(n^d)$ forward simulations.
In our default coarse-to-fine setting, this stage used 250 forward simulations.

The local stage is more expensive per iteration because it performs forward--backward differentiable MPM simulation~\cite{YHuICLR2020} over the full observed sequence.
We used 2,000 Lagrangian particle refinement iterations in the default setting.
In a representative timing measurement, the global search accounted for approximately 3\% of the local-refinement runtime because each global evaluation is forward-only and uses fewer frames.
We provide detailed runtime analysis in Appendix~\ref{subsec:runtime} and optimizer ablations in Appendix~\ref{subsec:global_optimizer_ablation}.

\section{Experiments}
\label{sec:experiments}

\subsection{Experimental Setup}
\label{subsec:experimental_setup}

\noindent\textbf{Datasets.}
We evaluated our method on the standard SfC benchmark~\cite{TKanekoCVPR2025} and the Topology dataset.
The SfC benchmark contains 40 sequences over five external shapes (\textit{sphere}, \textit{cube}, \textit{bicone}, \textit{cylinder}, and \textit{diamond}) with four cavity sizes and four cavity positions.
The Topology dataset is designed as an interior-topology stress test with 60 configurations grouped into \textit{symmetric}, \textit{asymmetric}, and \textit{complex} subsets.
Detailed dataset construction and representative examples are provided in Appendix~\ref{subsec:dataset}.

\smallskip
\noindent\textbf{Baselines.}
Our primary baseline is SfC-NeRF~\cite{TKanekoCVPR2025}, a density-based method that optimizes volumetric densities through differentiable physics simulation~\cite{YHuICLR2020}.
For future prediction, we also compared against GIC~\cite{JCaiNeurIPS2024}, which estimates physical parameters under a fixed dense interior, to test the benefit of explicit interior reconstruction.

\smallskip
\noindent\textbf{Metrics.}
For reconstruction, we reported the CD and F-score~\cite{MTatarchenkoCVPR2019} between the reconstructed and ground-truth particle configurations.
CD measures the geometric proximity, while the F-score evaluates the thresholded reconstruction consistency.
We set the F-score threshold to 0.5\% of the maximum bounding-box length and scale CD by $10^3$ for readability.
For future prediction, we reported the peak signal-to-noise ratio (PSNR) and structural similarity index (SSIM)~\cite{ZWangTIP2004} for rendered future frames.

\smallskip
\noindent\textbf{Default configuration.}
Unless otherwise stated, we followed the default SfC protocol~\cite{TKanekoCVPR2025}: the cavity size is known, with the Young's modulus $E=10^6$ and Poisson's ratio $\nu=0.3$.
Each sequence contains $N=14$ observed frames rendered from $M=11$ views around the collision event.
For our method, we used the known-size setting (i) in Section~\ref{subsec:hypothesis_space}, initializing the cavity from the observed external geometry without a ground-truth template.
This initialization closely matches centered cavities under size variation, while under position variation, cavities can differ in both position and geometry.
Hyperparameters are fixed across datasets, as described in Section~\ref{subsec:implementation}.

\subsection{Main Results on the SfC Benchmark}
\label{subsec:main_results}

We first evaluated hidden-structure reconstruction on the SfC benchmark, which varies the cavity size and position across five external shapes.

\smallskip
\noindent\textbf{Quantitative results.}
Table~\ref{tab:main_results} reports CD and F-score for the size- and position-variation subsets.
Our method outperforms SfC-NeRF in both settings, with larger gains under position variation, where cavity localization is more ambiguous.
This supports the failure analysis in Section~\ref{subsec:failure_analysis}. Density-based local optimization can be misled when surface-motion discrepancies do not directly indicate the location of hidden cavities.
By evaluating the explicit structural hypotheses through physical simulation, the global stage reduces this ambiguity before Lagrangian particle refinement.

\begin{table}[t]
  \centering
  \small
  \renewcommand{\arraystretch}{0.95}
  \setlength{\tabcolsep}{4.5pt}
  \begin{tabular}{lcccc}
    \toprule
    & \multicolumn{2}{c}{Size} & \multicolumn{2}{c}{Position} \\
    \cmidrule(lr){2-3}\cmidrule(lr){4-5}
    Method & CD$\downarrow$ & F$\uparrow$ & CD$\downarrow$ & F$\uparrow$ \\
    \midrule
    SfC-NeRF~\cite{TKanekoCVPR2025} & 0.165 & 0.135 & 0.281 & 0.128 \\
    Ours     & \textbf{0.063} & \textbf{0.151} & \textbf{0.109} & \textbf{0.144} \\
    \bottomrule
  \end{tabular}
  \vspace{-2mm}
  \caption{Main reconstruction results on the SfC benchmark.
    Chamfer distance (CD$\downarrow$) and F-score (F$\uparrow$) are reported for the cavity size and position variations.
    CD values are scaled by $10^3$.}
  \label{tab:main_results}
  \vspace{-1mm}
\end{table}

\begin{figure*}[t]
  \centering
  \includegraphics[width=0.97\textwidth]{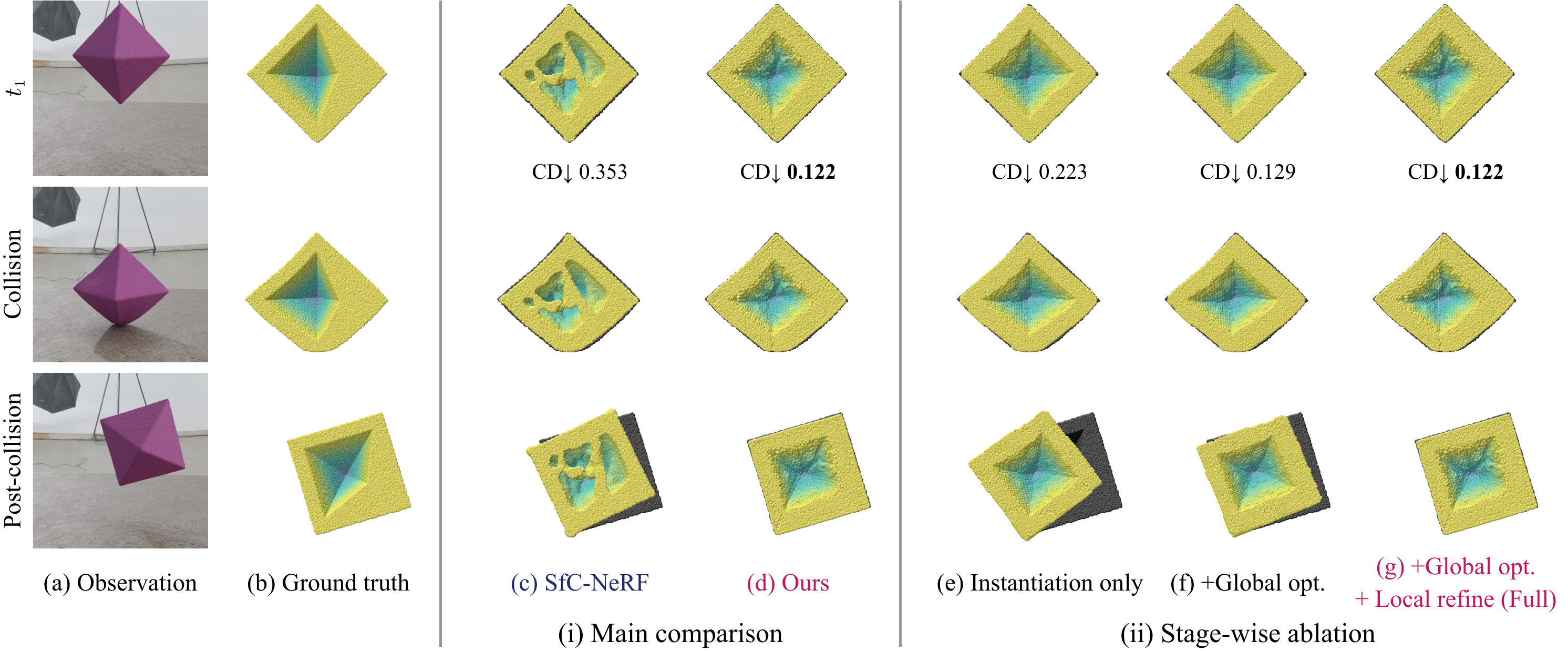}
  \vspace{-2mm}
  \caption{Qualitative results on the SfC benchmark under cavity position variation.
    (i) Main comparison. Our method reconstructs a more coherent internal cavity than SfC-NeRF and better matches the observed post-collision dynamics.
    (ii) Stage-wise ablation shows the effects of cavity instantiation, global hypothesis optimization, and Lagrangian particle refinement.
    The Chamfer distance (CD$\downarrow$) is reported below each reconstruction.
    Ground-truth cross-sections are shown in gray.}
  \label{fig:qualitative}
  \vspace{-4.5mm}
\end{figure*}

\smallskip
\noindent\textbf{Qualitative comparison.}
Fig.~\ref{fig:qualitative}(i) shows a representative example of cavity position variation.
SfC-NeRF reconstructs a less coherent internal cavity, which can lead to post-collision dynamics that deviate from the observation.
In contrast, our method reconstructs a more coherent cavity structure and better matches the observed sequence.
Additional qualitative results are available in Appendix~\ref{sec:qualitative} and on the \href{https://www.kecl.ntt.co.jp/people/kaneko.takuhiro/projects/sfc-hypothesis/}{project page}.

\subsection{Ablation Study}
\label{subsec:ablation}

We ablated the main components on the SfC benchmark under cavity position variation, where cavity localization is the most ambiguous.

\begin{table}[t]
  \centering
  \small
  \renewcommand{\arraystretch}{0.95}
  \setlength{\tabcolsep}{4.5pt}
  \begin{tabular}{lcc}
    \toprule
    Configuration & CD$\downarrow$ & F$\uparrow$ \\
    \midrule
    \multicolumn{3}{l}{\emph{(a) Framework stages}} \\
    Instantiation only                    & 0.204 & 0.136 \\
    +Local refine                         & 0.156 & 0.138 \\
    +Global opt.                          & 0.121 & 0.141 \\
    +Global opt. + Local refine (Full)    & \textbf{0.109} & \textbf{0.144} \\
    \midrule
    \multicolumn{3}{l}{\emph{(b) Geometry-aware losses for local refinement}} \\
    None                                  & 0.136 & 0.124 \\
    Rep. only                             & 0.124 & 0.142 \\
    Rep. + Fixed volume                   & 0.112 & 0.143 \\
    Rep. + Adaptive volume (Full)         & \textbf{0.109} & \textbf{0.144} \\
    \bottomrule
  \end{tabular}
  \vspace{-2mm}
  \caption{Ablation study on the SfC benchmark under cavity position variation.
    (a) Effect of framework stages.
    (b) Effect of geometry-aware losses during Lagrangian particle refinement.
    CD values are scaled by $10^3$.}
  \label{tab:ablation}
  \vspace{-4mm}
\end{table}

\smallskip
\noindent\textbf{Framework stages.}
Table~\ref{tab:ablation}(a) evaluates the framework stages, starting from the exterior-derived cavity instantiation, which encodes a structural prior.
Local refinement improves this initial configuration, while global hypothesis optimization yields a larger improvement by reducing ambiguity in cavity localization.
The full pipeline performs the best, indicating that the structural prior, global ambiguity reduction, and local geometric refinement play complementary roles.
Fig.~\ref{fig:qualitative}(ii) shows the same trend.

\smallskip
\noindent\textbf{Geometry-aware losses.}
Table~\ref{tab:ablation}(b) shows that both geometry-aware losses improve the Lagrangian refinement.
The repulsive loss stabilizes the particle spacing, and the volume-preservation loss further improves the reconstruction, with the adaptive variant performing the best.

\subsection{Interior-Topology Stress Tests}
\label{subsec:topology_stress}

We used the Topology dataset to stress-test reconstruction of more complex and ambiguous interior topologies than those in the standard SfC benchmark.
Table~\ref{tab:topology} shows how structural assumptions encoded in $\mathcal{H}$ affect reconstruction.
The candidate-template setting includes the ground-truth template and assumes a known center, whereas the known-template setting provides both the ground-truth template and center.
These controlled settings are stress tests rather than evidence of arbitrary topology discovery.

\begin{table}[t]
  \vspace{1mm}
  \centering
  \small
  \renewcommand{\arraystretch}{0.95}
  \setlength{\tabcolsep}{4.5pt}
  \begin{tabular}{@{}llccc@{}}
    \toprule
    Method & Setting & Sym & Asym & Comp \\
    \midrule
    SfC-NeRF~\cite{TKanekoCVPR2025}
           & Size
                     & 0.174 & 0.287 & 0.728 \\
    Ours
           & Size
                     & \textbf{0.142} & \textbf{0.185} & \textbf{0.582} \\
    \midrule
    Ours
           & Cand. w/ center
                     & 0.111 & 0.066 & 0.083 \\
    Ours
           & Template w/ center
                     & 0.090 & 0.066 & 0.070 \\
    \bottomrule
  \end{tabular}
  \vspace{-2mm}
  \caption{Interior-topology stress tests under different hypothesis-space settings.
    We report CD$\downarrow$ scaled by $10^3$ on symmetric (Sym), asymmetric (Asym), and complex (Comp) subsets.
    \emph{Size} denotes the known-size setting, \emph{Cand. w/ center} uses a set of candidate templates and a known center, and \emph{Template w/ center} uses a known template and center.
    The upper rows compare methods under the known-size setting, while the lower rows vary structural assumptions.
    F-score results are reported in Appendix~\ref{subsubsec:topology_fscore}.}
  \label{tab:topology}
  \vspace{-5mm}
\end{table}

\smallskip
\noindent\textbf{Known-size comparison.}
Under the known-size setting, our method outperforms SfC-NeRF across all topology subsets.
The error remains relatively high for Comp, indicating that the detailed reconstruction of complex cavities from collision observations alone is still difficult.

\smallskip
\noindent\textbf{Effect of structural assumptions.}
The lower rows show how changing $\mathcal{H}$ affects the result.
A set of candidate templates with a known center enables the global stage to compare structural alternatives before refinement, while a known template and center leave orientation as the main ambiguity.
These rows are not directly comparable because the settings provide different amounts of structural information.
Rather, they show how the same inference framework incorporates different structural assumptions through the hypothesis space.

\subsection{Future Prediction}
\label{subsec:future_prediction}

A recovered interior structure should support predictive simulation.
We evaluated future prediction on the SfC benchmark under cavity position variation, where incorrect cavity localization strongly affects post-collision dynamics.
Each method used $N=14$ observed frames and was evaluated at $+7$ and $+14$ future frames.
GIC and both variants of our method used the same GIC-based rendering pipeline, including the re-optimization of Gaussian attributes described in Appendix~\ref{subsubsec:relearning}, while SfC-NeRF used its original PAC-NeRF-based rendering pipeline.
Additional results on the SfC and Topology datasets are provided in Appendix~\ref{subsec:additional_future_prediction}.

\begin{table}[t]
  \centering
  \small
  \renewcommand{\arraystretch}{0.95}
  \setlength{\tabcolsep}{4.5pt}
  \begin{tabular}{lcccc}
    \toprule
    & \multicolumn{2}{c}{$+7$ frames} & \multicolumn{2}{c}{$+14$ frames} \\
    \cmidrule(lr){2-3}\cmidrule(lr){4-5}
    Method & PSNR$\uparrow$ & SSIM$\uparrow$ & PSNR$\uparrow$ & SSIM$\uparrow$ \\
    \midrule
    GIC~\cite{JCaiNeurIPS2024}        & 23.1 & 0.977 & 22.1 & 0.974 \\
    SfC-NeRF~\cite{TKanekoCVPR2025}   & 27.1 & 0.982 & 25.2 & 0.978 \\
    \midrule
    Ours (Global only)                & 24.9 & 0.980 & 23.6 & 0.977 \\
    Ours (Full)                       & \textbf{33.6} & \textbf{0.991} & \textbf{30.8} & \textbf{0.988} \\
    \bottomrule
  \end{tabular}
  \vspace{-2mm}
  \caption{Future prediction on the SfC benchmark under cavity position variation.
    PSNR and SSIM are reported at $+7$ and $+14$ frames beyond the observed window of $N=14$ frames.}
  \label{tab:future}
  \vspace{-1mm}
\end{table}

\smallskip
\noindent\textbf{Comparison with baselines.}
Table~\ref{tab:future} shows that our full method outperforms GIC and SfC-NeRF at both prediction horizons, particularly at the longer horizon.
The stronger long-horizon performance suggests a benefit from explicit hidden-structure reconstruction for predictive simulation.
We also note that the comparison with SfC-NeRF reflects differences in the rendering pipeline.

\smallskip
\noindent\textbf{Effect of local refinement.}
Lagrangian particle refinement corrects residual geometric discrepancies after the global stage and substantially improves prediction at both horizons.
Larger gains in future-frame metrics than in CD may arise because CD measures only the initial reconstructed geometry, while future-frame metrics also reflect the simulated trajectory and can be more sensitive to local geometric differences.
Together with the reconstruction ablation in Section~\ref{subsec:ablation}, these results support the complementary roles of the global and local stages.

\subsection{Robustness and Practical Considerations}
\label{subsec:robustness}

We evaluated the robustness of spherical objects from the SfC benchmark under cavity size and position variations, isolating internal-structure ambiguity from external-shape differences.
Table~\ref{tab:robustness} summarizes the representative results for optimizer changes, physical-parameter uncertainty, and degraded observations.

\begin{table}[t]
  \centering
  \small
  \renewcommand{\arraystretch}{0.95}
  \setlength{\tabcolsep}{4.5pt}
  \begin{tabular}{lcccc}
    \toprule
    & \multicolumn{2}{c}{Size} & \multicolumn{2}{c}{Position} \\
    \cmidrule(lr){2-3}\cmidrule(lr){4-5}
    Condition & CD$\downarrow$ & F$\uparrow$ & CD$\downarrow$ & F$\uparrow$ \\
    \midrule
    SfC-NeRF~\cite{TKanekoCVPR2025}     & 0.067 & 0.093 & 0.151 & 0.087 \\
    \midrule
    Ours (Clean)                        & 0.059 & 0.097 & 0.100 & 0.093 \\
    \multicolumn{5}{l}{\emph{(a) Optimizer}} \\
    Ours (Random search)                & 0.059 & 0.098 & 0.101 & 0.093 \\
    Ours (CMA-ES~\cite{NHansenEC2001})  & 0.059 & 0.097 & 0.102 & 0.093 \\
    \multicolumn{5}{l}{\emph{(b) Physical parameters}} \\
    Ours ($E$, $\pm 10\%$ error)        & 0.061 & 0.096 & 0.100 & 0.093 \\
    Ours ($\nu$, $\pm 10\%$ error)      & 0.059 & 0.097 & 0.101 & 0.093 \\
    Ours (Unknown phys. params.)        & 0.060 & 0.097 & 0.119 & 0.091 \\
    \multicolumn{5}{l}{\emph{(c) Observations}} \\
    Ours (Motion blur)                  & 0.064 & 0.095 & 0.096 & 0.093 \\
    Ours (Half FPS)                     & 0.061 & 0.096 & 0.112 & 0.091 \\
    \bottomrule
  \end{tabular}
  \vspace{-2mm}
  \caption{Robustness results on spherical objects under cavity size and position variations.
    (a) Optimizer variants use the same global objective and simulation budget as the default grid enumeration.
    (b) Physical-parameter errors are averaged over positive and negative perturbations; unknown physical parameters are jointly estimated.
    (c) Motion blur uses shutter angle $0.5$, and half FPS halves the observation frame rate.
    CD values are scaled by $10^3$.}
  \label{tab:robustness}
  \vspace{-4mm}
\end{table}

\smallskip
\noindent\textbf{Optimizer sensitivity.}
The random search and Covariance Matrix Adaptation Evolution Strategy (CMA-ES) match the default grid enumeration within 0.002 CD, with nearly identical F-scores, suggesting limited sensitivity to the choice of gradient-free optimizer.
Additional ablations are provided in Appendix~\ref{subsec:global_optimizer_ablation}.

\smallskip
\noindent\textbf{Toward less controlled settings.}
Moderate errors in $E$ or $\nu$ and motion blur have little effect.
Joint physical-parameter estimation remains close to the clean setting for size variation but causes moderate degradation for position variation.
Reducing the frame rate also moderately degrades performance in the position subset.
Additional robustness analyses are provided in Appendices~\ref{subsec:relaxing_assumptions}--\ref{subsec:material_variation}.

\section{Limitations and Discussion}
\label{sec:limitations}

This study has several limitations.
First, our experiments used controlled synthetic data with calibrated cameras and a known simulation model.
Although Section~\ref{subsec:robustness} evaluates motion blur and reduced frame rate, real-world capture can introduce additional errors, such as calibration noise and forward-model mismatch. The evaluation on in-the-wild collision sequences remains a topic for future work.

Second, the exact recovery of complex interior topology remains ill-posed.
As shown in Section~\ref{subsec:topology_stress}, collision observations leave residual ambiguities under weak structural assumptions, particularly for intricate geometries.
Hypothesis-space reasoning makes this trade-off explicit but cannot resolve insufficient observations or structures that are not adequately represented in the hypothesis space.
In the latter case, the global stage can only select the best available mismatched hypothesis, while Lagrangian particle refinement adjusts the geometry but preserves the cavity structure.
Thus, substantial topology changes must be represented in the hypothesis space before refinement.

Future work will include richer structural variables, such as multiple disconnected cavities, anisotropic scaling, and affine transformations, possibly combined with learned shape priors.
CMA-ES performs comparably to coarse-to-fine enumeration in the current low-dimensional settings, while adaptive sampling may become useful as the hypothesis space grows.
Extending the framework to interactions such as squeezing or shaking may also support internal-structure inference from everyday object dynamics.

\section{Conclusion}
\label{sec:conclusion}

We presented SfC as hypothesis-space reasoning, replacing implicit density updates with explicit structural hypotheses.
Our two-stage framework combines global hypothesis optimization and Lagrangian particle refinement, separating coarse structural reasoning from fine geometric adjustment.
Experiments showed improved reconstruction and future prediction over the density-based SfC baseline, while topology stress tests and robustness analyses clarified its scope and limitations.
This framework supports the structured inference of the hidden interior structure from collision-induced dynamics.

{
  \small
  \bibliographystyle{ieeenat_fullname}
  \bibliography{refs}

\begin{thebibliography}{52}
\providecommand{\natexlab}[1]{#1}
\providecommand{\url}[1]{\texttt{#1}}
\expandafter\ifx\csname urlstyle\endcsname\relax
  \providecommand{\doi}[1]{doi: #1}\else
  \providecommand{\doi}{doi: \begingroup \urlstyle{rm}\Url}\fi

\bibitem[Abou-Chakra et~al.(2024{\natexlab{a}})Abou-Chakra, Dayoub, and
  S{\"u}nderhauf]{JAbouWACV2024}
Jad Abou-Chakra, Feras Dayoub, and Niko S{\"u}nderhauf.
\newblock {ParticleNeRF}: A particle-based encoding for online neural radiance
  fields.
\newblock In \emph{WACV}, 2024{\natexlab{a}}.

\bibitem[Abou-Chakra et~al.(2024{\natexlab{b}})Abou-Chakra, Rana, Dayoub, and
  S{\"u}nderhauf]{JAbouCoRL2024}
Jad Abou-Chakra, Krishan Rana, Feras Dayoub, and Niko S{\"u}nderhauf.
\newblock Physically embodied {Gaussian} splatting: A visually learnt and
  physically grounded {3D} representation for robotics.
\newblock In \emph{CoRL}, 2024{\natexlab{b}}.

\bibitem[Cai et~al.(2024)Cai, Yang, Yuan, He, Dong, Bo, Cheng, and
  Chen]{JCaiNeurIPS2024}
Junhao Cai, Yuji Yang, Weihao Yuan, Yisheng He, Zilong Dong, Liefeng Bo, Hui
  Cheng, and Qifeng Chen.
\newblock {GIC}: {Gaussian}-informed continuum for physical property
  identification and simulation.
\newblock In \emph{NeurIPS}, 2024.

\bibitem[Chu et~al.(2022)Chu, Liu, Zheng, Franz, Seidel, Theobalt, and
  Zayer]{MChuTOG2022}
Mengyu Chu, Lingjie Liu, Quan Zheng, Erik Franz, Hans-Peter Seidel, Christian
  Theobalt, and Rhaleb Zayer.
\newblock Physics informed neural fields for smoke reconstruction with sparse
  data.
\newblock \emph{ACM Trans. Graph.}, 41\penalty0 (4), 2022.

\bibitem[Feng et~al.(2024)Feng, Shang, Li, Shao, Jiang, and
  Yang]{YFengCVPR2024}
Yutao Feng, Yintong Shang, Xuan Li, Tianjia Shao, Chenfanfu Jiang, and Yin
  Yang.
\newblock {PIE-NeRF}: Physics-based interactive elastodynamics with {NeRF}.
\newblock In \emph{CVPR}, 2024.

\bibitem[Furukawa and Ponce(2009)]{YFurukawaTPAMI2009}
Yasutaka Furukawa and Jean Ponce.
\newblock Accurate, dense, and robust multiview stereopsis.
\newblock \emph{IEEE Trans. Pattern Anal. Mach. Intell.}, 32\penalty0
  (8):\penalty0 1362--1376, 2009.

\bibitem[Gafni et~al.(2021)Gafni, Thies, Zollhofer, and
  Nie{\ss}ner]{GGafniCVPR2021}
Guy Gafni, Justus Thies, Michael Zollhofer, and Matthias Nie{\ss}ner.
\newblock Dynamic neural radiance fields for monocular {4D} facial avatar
  reconstruction.
\newblock In \emph{CVPR}, 2021.

\bibitem[Guan et~al.(2022)Guan, Deng, Wang, and Yang]{SGuanICML2022}
Shanyan Guan, Huayu Deng, Yunbo Wang, and Xiaokang Yang.
\newblock {NeuroFluid}: Fluid dynamics grounding with particle-driven neural
  radiance fields.
\newblock In \emph{ICML}, 2022.

\bibitem[Hansen and Ostermeier(2001)]{NHansenEC2001}
Nikolaus Hansen and Andreas Ostermeier.
\newblock Completely derandomized self-adaptation in evolution strategies.
\newblock \emph{Evol. Comput.}, 9\penalty0 (2):\penalty0 159--195, 2001.

\bibitem[Hu et~al.(2018)Hu, Fang, Ge, Qu, Zhu, Pradhana, and Jiang]{YHuTOG2018}
Yuanming Hu, Yu Fang, Ziheng Ge, Ziyin Qu, Yixin Zhu, Andre Pradhana, and
  Chenfanfu Jiang.
\newblock A moving least squares material point method with displacement
  discontinuity and two-way rigid body coupling.
\newblock \emph{ACM Trans. Graph.}, 37\penalty0 (4), 2018.

\bibitem[Hu et~al.(2020)Hu, Anderson, Li, Sun, Carr, Ragan-Kelley, and
  Durand]{YHuICLR2020}
Yuanming Hu, Luke Anderson, Tzu-Mao Li, Qi Sun, Nathan Carr, Jonathan
  Ragan-Kelley, and Fr{\'e}do Durand.
\newblock {DiffTaichi}: Differentiable programming for physical simulation.
\newblock In \emph{ICLR}, 2020.

\bibitem[Huang et~al.(2025)Huang, Zhang, Zeng, Zhang, Li, Zuo, and
  Lau]{THuangAAAI2025}
Tianyu Huang, Haoze Zhang, Yihan Zeng, Zhilu Zhang, Hui Li, Wangmeng Zuo, and
  Rynson W.~H. Lau.
\newblock {DreamPhysics}: Learning physics-based {3D} dynamics with video
  diffusion priors.
\newblock In \emph{AAAI}, 2025.

\bibitem[Jiang et~al.(2025)Jiang, Hsu, Zhang, Yu, Wang, and Li]{HJiangICCV2025}
Hanxiao Jiang, Hao-Yu Hsu, Kaifeng Zhang, Hsin-Ni Yu, Shenlong Wang, and Yunzhu
  Li.
\newblock {PhysTwin}: Physics-informed reconstruction and simulation of
  deformable objects from videos.
\newblock In \emph{ICCV}, 2025.

\bibitem[Jiang et~al.(2024)Jiang, Yu, Xie, Li, Feng, Wang, Li, Lau, Gao, Yang,
  and Jiang]{YJiangTOG2024}
Ying Jiang, Chang Yu, Tianyi Xie, Xuan Li, Yutao Feng, Huamin Wang, Minchen Li,
  Henry Lau, Feng Gao, Yin Yang, and Chenfanfu Jiang.
\newblock {VR-GS}: A physical dynamics-aware interactive {Gaussian} splatting
  system in virtual reality.
\newblock In \emph{SIGGRAPH}, 2024.

\bibitem[Kaneko(2024)]{TKanekoCVPR2024}
Takuhiro Kaneko.
\newblock Improving physics-augmented continuum neural radiance field-based
  geometry-agnostic system identification with {Lagrangian} particle
  optimization.
\newblock In \emph{CVPR}, 2024.

\bibitem[Kaneko(2025)]{TKanekoCVPR2025}
Takuhiro Kaneko.
\newblock Structure from collision.
\newblock In \emph{CVPR}, 2025.

\bibitem[Katsumata et~al.(2024)Katsumata, Vo, and Nakayama]{KKatsumataECCV2024}
Kai Katsumata, Duc~Minh Vo, and Hideki Nakayama.
\newblock A compact dynamic {3D} {Gaussian} representation for real-time
  dynamic view synthesis.
\newblock In \emph{ECCV}, 2024.

\bibitem[Kerbl et~al.(2023)Kerbl, Kopanas, Leimk{\"u}hler, and
  Drettakis]{BKerblTOG2023}
Bernhard Kerbl, Georgios Kopanas, Thomas Leimk{\"u}hler, and George Drettakis.
\newblock {3D Gaussian} splatting for real-time radiance field rendering.
\newblock \emph{ACM Trans. Graph.}, 42\penalty0 (4), 2023.

\bibitem[Kingma and Ba(2015)]{DPKingmaICLR2015}
Diederik~P. Kingma and Jimmy Ba.
\newblock Adam: A method for stochastic optimization.
\newblock In \emph{ICLR}, 2015.

\bibitem[Li et~al.(2023)Li, Qiao, Chen, Jatavallabhula, Lin, Jiang, and
  Gan]{XLiICLR2023}
Xuan Li, Yi-Ling Qiao, Peter~Yichen Chen, Krishna~Murthy Jatavallabhula, Ming
  Lin, Chenfanfu Jiang, and Chuang Gan.
\newblock {PAC-NeRF}: Physics augmented continuum neural radiance fields for
  geometry-agnostic system identification.
\newblock In \emph{ICLR}, 2023.

\bibitem[Li et~al.(2021)Li, Niklaus, Snavely, and Wang]{ZLiCVPR2021}
Zhengqi Li, Simon Niklaus, Noah Snavely, and Oliver Wang.
\newblock Neural scene flow fields for space-time view synthesis of dynamic
  scenes.
\newblock In \emph{CVPR}, 2021.

\bibitem[Lin et~al.(2021)Lin, Ryabtsev, Sengupta, Curless, Seitz, and
  Kemelmacher-Shlizerman]{SLinCVPR2021}
Shanchuan Lin, Andrey Ryabtsev, Soumyadip Sengupta, Brian~L. Curless, Steven~M.
  Seitz, and Ira Kemelmacher-Shlizerman.
\newblock Real-time high-resolution background matting.
\newblock In \emph{CVPR}, 2021.

\bibitem[Liu et~al.(2026)Liu, Wang, Lin, Xiao, Chen, Wen, Zhang, Qi, Yang,
  Jeni, Xu, and Zhao]{CLiuICLR2026}
Chunjiang Liu, Xiaoyuan Wang, Qingran Lin, Albert Xiao, Haoyu Chen, Shizheng
  Wen, Hao Zhang, Lu Qi, Ming-Hsuan Yang, Laszlo~A. Jeni, Min Xu, and Yizhou
  Zhao.
\newblock Multi-object system identification from videos.
\newblock In \emph{ICLR}, 2026.

\bibitem[Liu et~al.(2024)Liu, Wang, Yao, Zhang, Zhou, and Duan]{FLiuArXiv2024}
Fangfu Liu, Hanyang Wang, Shunyu Yao, Shengjun Zhang, Jie Zhou, and Yueqi Duan.
\newblock {Physics3D}: Learning physical properties of {3D} {Gaussians} via
  video diffusion.
\newblock \emph{arXiv preprint arXiv:2406.04338}, 2024.

\bibitem[Liu et~al.(2025)Liu, Ye, Luximon, Wan, and Zhang]{ZLiuCVPR2025}
Zhuoman Liu, Weicai Ye, Yan Luximon, Pengfei Wan, and Di Zhang.
\newblock Unleashing the potential of multi-modal foundation models and video
  diffusion for {4D} dynamic physical scene simulation.
\newblock In \emph{CVPR}, 2025.

\bibitem[Lu et~al.(2024)Lu, Guo, Hui, Chen, Yang, Tang, Zhu, and
  Dai]{ZLuCVPR2024}
Zhicheng Lu, Xiang Guo, Le Hui, Tianrui Chen, Min Yang, Xiao Tang, Feng Zhu,
  and Yuchao Dai.
\newblock {3D} geometry-aware deformable {Gaussian} splatting for dynamic view
  synthesis.
\newblock In \emph{CVPR}, 2024.

\bibitem[Luiten et~al.(2024)Luiten, Kopanas, Leibe, and
  Ramanan]{JLuiten3DV2024}
Jonathon Luiten, Georgios Kopanas, Bastian Leibe, and Deva Ramanan.
\newblock Dynamic {3D Gaussians}: Tracking by persistent dynamic view
  synthesis.
\newblock In \emph{3DV}, 2024.

\bibitem[Mildenhall et~al.(2020)Mildenhall, Srinivasan, Tancik, Barron,
  Ramamoorthi, and Ng]{BMildenhallECCV2020}
Ben Mildenhall, Pratul~P. Srinivasan, Matthew Tancik, Jonathan~T. Barron, Ravi
  Ramamoorthi, and Ren Ng.
\newblock {NeRF}: Representing scenes as neural radiance fields for view
  synthesis.
\newblock In \emph{ECCV}, 2020.

\bibitem[Ni et~al.(2024)Ni, Chen, Jing, Jiang, Wang, Dai, Li, Zhu, Zhu, and
  Huang]{JNiNeurIPS2024}
Junfeng Ni, Yixin Chen, Bohan Jing, Nan Jiang, Bin Wang, Bo Dai, Puhao Li,
  Yixin Zhu, Song-Chun Zhu, and Siyuan Huang.
\newblock {PhyRecon}: Physically plausible neural scene reconstruction.
\newblock In \emph{NeurIPS}, 2024.

\bibitem[Park et~al.(2021)Park, Sinha, Barron, Bouaziz, Goldman, Seitz, and
  Martin-Brualla]{KParkICCV2021}
Keunhong Park, Utkarsh Sinha, Jonathan~T. Barron, Sofien Bouaziz, Dan~B.
  Goldman, Steven~M. Seitz, and Ricardo Martin-Brualla.
\newblock Nerfies: Deformable neural radiance fields.
\newblock In \emph{ICCV}, 2021.

\bibitem[Pumarola et~al.(2021)Pumarola, Corona, Pons-Moll, and
  Moreno-Noguer]{APumarolaCVPR2021}
Albert Pumarola, Enric Corona, Gerard Pons-Moll, and Francesc Moreno-Noguer.
\newblock {D-NeRF}: Neural radiance fields for dynamic scenes.
\newblock In \emph{CVPR}, 2021.

\bibitem[Qiu et~al.(2024)Qiu, Yang, Zeng, and Wang]{RZQiuECCV2024}
Ri-Zhao Qiu, Ge Yang, Weijia Zeng, and Xiaolong Wang.
\newblock {Feature Splatting}: Language-driven physics-based scene synthesis
  and editing.
\newblock In \emph{ECCV}, 2024.

\bibitem[Ravi et~al.(2025)Ravi, Gabeur, Hu, Hu, Ryali, Ma, Khedr, R{\"a}dle,
  Rolland, Gustafson, Mintun, Pan, Alwala, Carion, Wu, Girshick, Doll{\'a}r,
  and Feichtenhofer]{NRaviICLR2025}
Nikhila Ravi, Valentin Gabeur, Yuan-Ting Hu, Ronghang Hu, Chaitanya Ryali,
  Tengyu Ma, Haitham Khedr, Roman R{\"a}dle, Chloe Rolland, Laura Gustafson,
  Eric Mintun, Junting Pan, Kalyan~Vasudev Alwala, Nicolas Carion, Chao-Yuan
  Wu, Ross Girshick, Piotr Doll{\'a}r, and Christoph Feichtenhofer.
\newblock {SAM 2}: Segment anything in images and videos.
\newblock In \emph{ICLR}, 2025.

\bibitem[Seitz et~al.(2006)Seitz, Curless, Diebel, Scharstein, and
  Szeliski]{SSeitzCVPR2006}
Steven~M. Seitz, Brian Curless, James Diebel, Daniel Scharstein, and Richard
  Szeliski.
\newblock A comparison and evaluation of multi-view stereo reconstruction
  algorithms.
\newblock In \emph{CVPR}, 2006.

\bibitem[Shaikh et~al.(2026)Shaikh, Banerjee, and Gupta]{FShaikhCVPR2026}
Farhat Shaikh, Ayan Banerjee, and Sandeep Gupta.
\newblock {EMMA}: Extracting multiple physical parameters from multimodal data.
\newblock In \emph{CVPR}, 2026.

\bibitem[Snavely et~al.(2006)Snavely, Seitz, and Szeliski]{NSnavelyTOG2006}
Noah Snavely, Steven~M. Seitz, and Richard Szeliski.
\newblock Photo tourism: Exploring photo collections in {3D}.
\newblock \emph{ACM Trans. Graph.}, 25\penalty0 (3):\penalty0 835--846, 2006.

\bibitem[Sulsky et~al.(1994)Sulsky, Chen, and Schreyer]{DSulskyCMAME1994}
Deborah Sulsky, Zhen Chen, and Howard~L. Schreyer.
\newblock A particle method for history-dependent materials.
\newblock \emph{Comput. Methods Appl. Mech. Eng.}, 118\penalty0 (1-2):\penalty0
  179--196, 1994.

\bibitem[Sun et~al.(2022)Sun, Sun, and Chen]{CSunCVPR2022}
Cheng Sun, Min Sun, and Hwann-Tzong Chen.
\newblock Direct voxel grid optimization: Super-fast convergence for radiance
  fields reconstruction.
\newblock In \emph{CVPR}, 2022.

\bibitem[Tatarchenko et~al.(2019)Tatarchenko, Richter, Ranftl, Li, Koltun, and
  Brox]{MTatarchenkoCVPR2019}
Maxim Tatarchenko, Stephan~R. Richter, Ren{\'e} Ranftl, Zhuwen Li, Vladlen
  Koltun, and Thomas Brox.
\newblock What do single-view {3D} reconstruction networks learn?
\newblock In \emph{CVPR}, 2019.

\bibitem[Tomasi and Kanade(1992)]{CTomasiIJCV1992}
Carlo Tomasi and Takeo Kanade.
\newblock Shape and motion from image streams under orthography: a
  factorization method.
\newblock \emph{Int. J. Comput. Vis.}, 9\penalty0 (2):\penalty0 137--154, 1992.

\bibitem[Tretschk et~al.(2021)Tretschk, Tewari, Golyanik, Zollh{\"o}fer,
  Lassner, and Theobalt]{ETretschkICCV2021}
Edgar Tretschk, Ayush Tewari, Vladislav Golyanik, Michael Zollh{\"o}fer,
  Christoph Lassner, and Christian Theobalt.
\newblock Non-rigid neural radiance fields: Reconstruction and novel view
  synthesis of a dynamic scene from monocular video.
\newblock In \emph{ICCV}, 2021.

\bibitem[Vasile et~al.(2025)Vasile, Qiu, Natale, and Wang]{FVasileNeurIPS2025}
Federico Vasile, Ri-Zhao Qiu, Lorenzo Natale, and Xiaolong Wang.
\newblock {Gaussian}-augmented physics simulation and system identification
  with complex colliders.
\newblock In \emph{NeurIPS}, 2025.

\bibitem[Wada et~al.(2026)Wada, Yamashita, Kawahara, and
  Nishino]{NWadaCVPR2026}
Norika Wada, Kohei Yamashita, Ryo Kawahara, and Ko Nishino.
\newblock {M-PhyGs}: Multi-material object dynamics from video.
\newblock In \emph{CVPR Findings}, 2026.

\bibitem[Wang et~al.(2004)Wang, Bovik, Sheikh, and Simoncelli]{ZWangTIP2004}
Zhou Wang, Alan~C. Bovik, Hamid~R. Sheikh, and Eero~P. Simoncelli.
\newblock Image quality assessment: From error visibility to structural
  similarity.
\newblock \emph{IEEE Trans. Image Process.}, 13\penalty0 (4), 2004.

\bibitem[Wu et~al.(2024)Wu, Yi, Fang, Xie, Zhang, Wei, Liu, Tian, and
  Wang]{GWuCVPR2024}
Guanjun Wu, Taoran Yi, Jiemin Fang, Lingxi Xie, Xiaopeng Zhang, Wei Wei, Wenyu
  Liu, Qi Tian, and Xinggang Wang.
\newblock {4D} {Gaussian} splatting for real-time dynamic scene rendering.
\newblock In \emph{CVPR}, 2024.

\bibitem[Xie et~al.(2024)Xie, Zong, Qiu, Li, Feng, Yang, and
  Jiang]{TXieCVPR2024}
Tianyi Xie, Zeshun Zong, Yuxing Qiu, Xuan Li, Yutao Feng, Yin Yang, and
  Chenfanfu Jiang.
\newblock {PhysGaussian}: Physics-integrated {3D Gaussians} for generative
  dynamics.
\newblock In \emph{CVPR}, 2024.

\bibitem[Yan et~al.(2025)Yan, Peng, Wang, Tang, Yang, Liang, Wu, and
  Wang]{JYanCVPR2025}
Jinbo Yan, Rui Peng, Zhiyan Wang, Luyang Tang, Jiayu Yang, Jie Liang, Jiahao
  Wu, and Ronggang Wang.
\newblock Instant {Gaussian} stream: Fast and generalizable streaming of
  dynamic scene reconstruction via {Gaussian} splatting.
\newblock In \emph{CVPR}, 2025.

\bibitem[Yang et~al.(2024{\natexlab{a}})Yang, Gao, Zhou, Jiao, Zhang, and
  Jin]{ZYangCVPR2024}
Ziyi Yang, Xinyu Gao, Wen Zhou, Shaohui Jiao, Yuqing Zhang, and Xiaogang Jin.
\newblock Deformable {3D Gaussians} for high-fidelity monocular dynamic scene
  reconstruction.
\newblock In \emph{CVPR}, 2024{\natexlab{a}}.

\bibitem[Yang et~al.(2024{\natexlab{b}})Yang, Yang, Pan, and
  Zhang]{ZYangICLR2024}
Zeyu Yang, Hongye Yang, Zijie Pan, and Li Zhang.
\newblock Real-time photorealistic dynamic scene representation and rendering
  with {4D Gaussian} splatting.
\newblock In \emph{ICLR}, 2024{\natexlab{b}}.

\bibitem[Zhang et~al.(2024)Zhang, Yu, Wu, Feng, Zheng, Snavely, Wu, and
  Freeman]{TZhangECCV2024}
Tianyuan Zhang, Hong-Xing Yu, Rundi Wu, Brandon~Y. Feng, Changxi Zheng, Noah
  Snavely, Jiajun Wu, and William~T. Freeman.
\newblock {PhysDreamer}: Physics-based interaction with {3D} objects via video
  generation.
\newblock In \emph{ECCV}, 2024.

\bibitem[Zhao et~al.(2025)Zhao, Chen, Liu, Li, Herrmann, Hur, Li, Yang, Raj,
  and Xu]{YZhaoICCV2025}
Yizhou Zhao, Haoyu Chen, Chunjiang Liu, Zhenyang Li, Charles Herrmann, Junhwa
  Hur, Yinxiao Li, Ming-Hsuan Yang, Bhiksha Raj, and Min Xu.
\newblock Toward material-agnostic system identification from videos.
\newblock In \emph{ICCV}, 2025.

\bibitem[Zhong et~al.(2024)Zhong, Yu, Wu, and Li]{LZhongECCV2024}
Licheng Zhong, Hong-Xing Yu, Jiajun Wu, and Yunzhu Li.
\newblock Reconstruction and simulation of elastic objects with spring-mass
  {3D} {Gaussians}.
\newblock In \emph{ECCV}, 2024.

\end{thebibliography}
}

\clearpage
\addtocontents{toc}{\protect\setcounter{tocdepth}{2}}
\appendix

\begingroup
\renewcommand{\contentsname}{Supplementary Material}
\setcounter{tocdepth}{2}
\hypersetup{
  linkcolor=black,
  linktoc=all
}
\tableofcontents
\endgroup

\section{Implementation Details}
\label{sec:implementation}

\subsection{Dataset}
\label{subsec:dataset}

\subsubsection{SfC Dataset}
\label{subsubsec:sfc_dataset}

\begin{figure*}[htbp]
  \centering
  \includegraphics[width=0.99\textwidth]{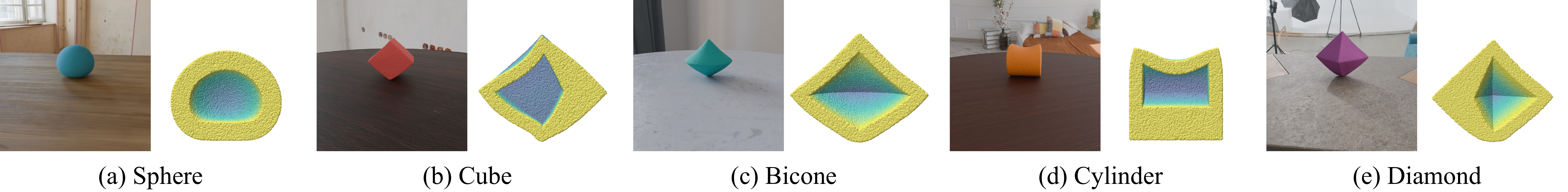}
  \vspace{-1mm}
  \caption{Overview of the SfC dataset.
    For each object, we show a representative observation image and the corresponding ground-truth cross-section.
    The dataset contains five external shapes ((a)~\textit{sphere}, (b)~\textit{cube}, (c)~\textit{bicone}, (d)~\textit{cylinder}, and (e)~\textit{diamond}) with cavity size and position variations.}
  \label{fig:dataset_sfc}
\end{figure*}

The SfC dataset~\cite{TKanekoCVPR2025} contains five external shapes: \textit{sphere}, \textit{cube}, \textit{bicone}, \textit{cylinder}, and \textit{diamond}.
The dataset includes two structural variation subsets used in our main experiments: \textit{cavity size} and \textit{cavity position}.
In the size variation subset, cavity sizes vary over
$s_c \in \left\{ 0, \left(\frac{1}{2}\right)^3, \left(\frac{2}{3}\right)^3, \left(\frac{3}{4}\right)^3 \right\}$
with the cavity center held fixed.
In the position variation subset, cavity locations vary over
$l_c \in \{\text{left}, \text{right}, \text{up}, \text{down}\}$
with the cavity size fixed to $s_c = \left(\frac{2}{3}\right)^3$.
All objects are composed of elastic material with Young's modulus $E=10^6$ and Poisson's ratio $\nu=0.3$, following the original SfC simulation setting.
In total, the structural benchmark comprises $5$ external shapes $\times$ $(4$ size settings $+$ $4$ position settings$) = 40$ sequences.
Representative examples are shown in Fig.~\ref{fig:dataset_sfc}.

Ground-truth dynamics are simulated using the Moving Least Squares Material Point Method (MLS-MPM) simulator~\cite{YHuTOG2018}, in which each object is dropped under gravity and allowed to collide with a ground plane.
Each sequence contains $14$ observed frames.
For each deformation state, multiview images are rendered from $11$ viewpoints arranged on the upper hemisphere surrounding the object.
Images are rendered with diverse environmental lighting conditions and ground textures to increase visual diversity.

\subsubsection{Topology Dataset}
\label{subsubsec:topology_dataset}

\begin{figure*}[htbp]
  \centering
  \begin{subfigure}{0.99\textwidth}
    \centering
    \includegraphics[width=0.99\textwidth]{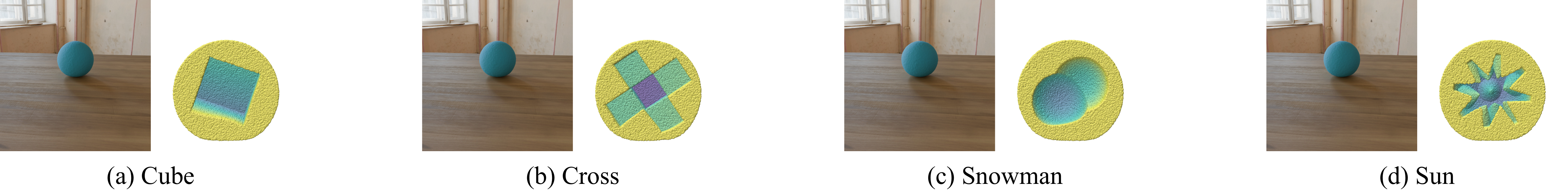}
    \caption{Symmetric}
    \vspace{2mm}
  \end{subfigure}
  \begin{subfigure}{0.99\textwidth}
    \centering
    \includegraphics[width=0.99\textwidth]{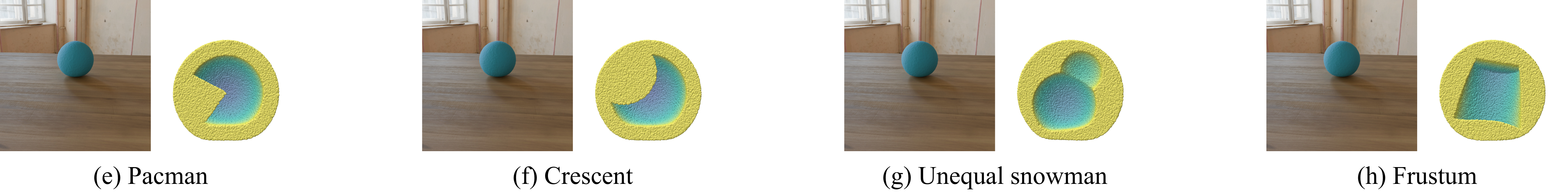}
    \caption{Asymmetric}
    \vspace{2mm}
  \end{subfigure}
  \begin{subfigure}{0.99\textwidth}
    \centering
    \includegraphics[width=0.99\textwidth]{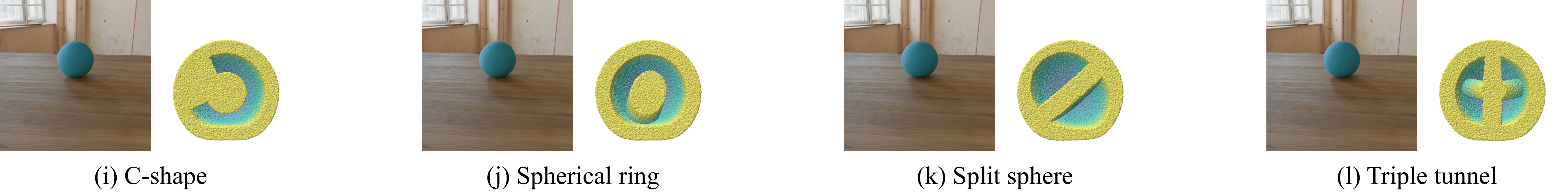}
    \caption{Complex}
    \vspace{-2mm}
  \end{subfigure}
  \caption{Overview of the Topology dataset.
    For each shape, we show a representative observation image and the corresponding ground-truth cross-section.
    The dataset contains three subsets:
    (I)~\textit{symmetric} ((a)~\textit{cube}, (b)~\textit{cross}, (c)~\textit{snowman}, (d)~\textit{sun}),
    (II)~\textit{asymmetric} ((e)~\textit{pacman}, (f)~\textit{crescent}, (g)~\textit{unequal snowman}, (h)~\textit{frustum}),
    and (III)~\textit{complex} ((i)~\textit{C-shape}, (j)~\textit{spherical ring}, (k)~\textit{split sphere}, (l)~\textit{triple tunnel}).}
  \label{fig:dataset_topology}
\end{figure*}

We used the Topology dataset to stress-test reconstruction under more complex interior topologies.
Unlike the SfC dataset, which varies cavity size and position, this dataset varies the internal cavity geometry and topology.
It is designed to study how structural assumptions encoded in the hypothesis space affect reconstruction, rather than to evaluate arbitrary topology discovery.
The dataset contains three subsets---\textit{symmetric}, \textit{asymmetric}, and \textit{complex}---each consisting of four cavity shapes.
Representative examples are shown in Fig.~\ref{fig:dataset_topology}.

\smallskip
\noindent\textit{Symmetric set.}
The symmetric set contains cavity shapes with discrete rotational symmetry about the object center.
It includes \textit{cube}, \textit{cross}, \textit{snowman}, and \textit{sun}.
The \textit{cube} defines an axis-aligned rectangular cavity.
The \textit{cross} is composed of three mutually orthogonal bars.
The \textit{snowman} consists of two equal spheres arranged along a single axis, exhibiting uniaxial symmetry.
The \textit{sun} extends conical protrusions in 26 discrete directions from a central sphere.
These shapes provide controlled cases for evaluating the effect of symmetric cavity geometry on reconstruction.

\smallskip
\noindent\textit{Asymmetric set.}
The asymmetric set contains cavity shapes that break symmetry and introduce a preferred orientation.
It includes \textit{pacman}, \textit{crescent}, \textit{unequal snowman}, and \textit{frustum}.
The \textit{pacman} is derived from a sphere by removing an angular wedge, while the \textit{crescent} is obtained by subtracting a shifted overlapping sphere.
The \textit{unequal snowman} combines two spheres of different radii to form an elongated asymmetric structure, and the \textit{frustum} defines a truncated cone with a tapered cross-section.
These shapes test sensitivity to orientation cues in hidden cavity geometry.

\smallskip
\noindent\textit{Complex set.}
The complex set contains cavity shapes with non-trivial topology or spatial discontinuity.
It includes \textit{C-shape}, \textit{spherical ring}, \textit{split sphere}, and \textit{triple tunnel}.
The \textit{C-shape} forms a partially enclosed shell; the \textit{spherical ring} retains a cylindrical core within a spherical cavity; and the \textit{split sphere} produces two disjoint half-spherical cavities displaced along a single axis.
The \textit{triple tunnel} embeds three orthogonal cylindrical channels within a spherical cavity, yielding a genus-3 topology.
These shapes stress-test reconstruction under high curvature and topological complexity.

All configurations use a spherical exterior and fixed cavity size $s_c=\left(\tfrac{2}{3}\right)^3$.
We generated five random rotations for each shape, yielding $4$ shapes $\times$ $3$ subsets $\times$ $5$ rotations $=60$ configurations.
Simulation and rendering follow the SfC dataset settings, with $14$ observed frames and $11$ upper-hemisphere views per sequence.

\subsection{Model Details}
\label{subsec:model_details}

An overview of the proposed pipeline is shown in Fig.~\ref{fig:overview} of the main paper.
We describe the implementation details of each step below.

\subsubsection{Step (0). Dense Initialization}
\label{subsubsec:initialization}

We initialized the object with a fully dense particle representation using GIC~\cite{JCaiNeurIPS2024}.
GIC first reconstructs a dynamic 3D Gaussian representation from multiview observations and then converts it into a continuum particle representation for MPM simulation.
Each continuum particle is associated with a position, scale, and density value, which are used for simulation and mask rendering.
We used the official GIC implementation and followed its default training configuration.
Below, we describe the GIC-based initialization procedure, including dynamic 3D reconstruction, continuum generation, physical simulation, and velocity estimation.

\smallskip
\noindent\textbf{Dynamic 3D reconstruction.}
GIC reconstructs dynamic scenes using a canonical set of 3D Gaussians and a learned deformation field that maps the canonical configuration to each time step.
The model is trained with L1 and SSIM~\cite{ZWangTIP2004} losses, following the architecture and hyperparameters of the official GIC implementation.

\smallskip
\noindent\textbf{Continuum generation.}
From the reconstructed Gaussian representation, we generated a fully dense grid-based point set using a coarse-to-fine strategy.
The initial grid size was set to $d_{\mathrm{grid}}^0 = 0.16$, and the grid was upsampled four times.
This yielded a final grid spacing of $d_{\mathrm{grid}} = 0.16 / 2^4 = 0.01$.
This spacing matched the average inter-particle distance used in SfC-NeRF~\cite{TKanekoCVPR2025}, so that the comparison was not affected by particle density.
In practice, this resolution produced approximately $10^5$ particles for a fully dense object before cavity removal.

\smallskip
\noindent\textbf{Physical simulation.}
We used a differentiable MPM simulator implemented with DiffTaichi~\cite{YHuICLR2020}.
During simulation, object silhouettes are rendered as alpha masks using 3DGS~\cite{BKerblTOG2023}.
These masks are compared with the observed image masks to compute the 2D observation loss.
The simulation parameters, including the material model and time-step settings, follow those used in SfC-NeRF~\cite{TKanekoCVPR2025} for a fair comparison.

\smallskip
\noindent\textbf{Velocity estimation.}
We estimated the pre-collision falling velocity following the procedure used in GIC~\cite{JCaiNeurIPS2024}.
We used four pre-impact frames with the default hyperparameters from the official GIC implementation.

\subsubsection{Step (1). Hypothesis Instantiation}
\label{subsubsec:hypothesis_instantiation}

Starting from the dense particle initialization, we instantiated candidate cavity hypotheses using the structural parameterization described in Section~\ref{subsec:hypothesis_space} of the main paper.
Each hypothesis is specified by compact structural variables, including a cavity shape representation, center position, size, and orientation.
The cavity shape representation may be derived from the observed external geometry or selected from a set of templates, depending on the available structural assumptions.
We removed particles inside the resulting cavity region to form the corresponding internal structure.
Structural assumptions fix or restrict a subset of these variables, while the remaining variables are optimized in Step~(2).

\subsubsection{Step (2). Global Hypothesis Optimization}
\label{subsubsec:global_hypothesis_optimization}

The global stage evaluates candidate cavity hypotheses using forward-only MPM simulation~\cite{DSulskyCMAME1994,YHuTOG2018} and selects the lowest-loss hypothesis.

\smallskip
\noindent\textbf{Default optimizer.}
Unless otherwise stated, we used coarse-to-fine enumeration as the default gradient-free optimizer.
This choice is deterministic, reproducible, and consistent with the main experiments.

\smallskip
\noindent\textbf{Position-search regularization.}
For position search, we augmented $\mathcal{L}_{\mathrm{global}}(h)$ with the displacement regularizer
\begin{align}
  \mathcal{R}_{\mathrm{disp}}(h)
  =
  e^{-k_{\mathrm{disp}} r_h}
  \left\|
  \widetilde{\bm{\Delta}}_h
  \right\|_2^2,
  \label{eq:position_displacement_regularizer}
\end{align}
where $\widetilde{\bm{\Delta}}_h$ denotes the cavity displacement from its initial placement in voxel-index units, normalized element-wise by the index span of the object mask along each axis, with each span lower-bounded by one.
The cavity-volume ratio $r_h$ is the density volume inside the shifted cavity divided by that of the original object.
Hypotheses whose shifted cavity reaches the object boundary are not rendered and are treated as invalid during hypothesis selection.
We minimized $\mathcal{L}_{\mathrm{global}}(h)+w_{\mathrm{disp}}\mathcal{R}_{\mathrm{disp}}(h)$ with $k_{\mathrm{disp}}=5$ and $w_{\mathrm{disp}}=2$.
This regularizer discourages cavity displacements that are disproportionately large relative to the object span.
The penalty is measured from the initial cavity placement, with its weight decreasing as the cavity-volume ratio increases.

For the unknown-size experiment, we instead used $w_{\mathrm{disp}}=1$ and added $w_{\mathrm{size}}\mathcal{R}_{\mathrm{size}}(h)$ to the objective.
The size regularizer is
\begin{align}
  \mathcal{R}_{\mathrm{size}}(h)
  =
  \frac{
  e^{k_{\mathrm{size}}r_h}-1
  }{
  e^{k_{\mathrm{size}}}-1
  },
  \label{eq:position_size_regularizer}
\end{align}
with $k_{\mathrm{size}}=5$ and $w_{\mathrm{size}}=0.1$.
This regularizer penalizes large cavity-volume ratios and is normalized to $[0,1]$.

\smallskip
\noindent\textbf{Setting-specific search variables.}
The optimized variables and search ranges depend on the hypothesis-space setting, as detailed below.

\smallskip
\noindent\textit{(i) Known size.}
In this setting, we constructed the cavity by offsetting inward from the observed external surface without using a ground-truth template.
The resulting cavity shape and the specified size are fixed, and the default search optimizes only its position.
The default grid optimizer first performs a coarse $5 \times 5 \times 5$ position search over the three Cartesian axes of the object interior, followed by a refined $5 \times 5 \times 5$ search around the best candidate.
This evaluates $250$ hypotheses in total ($125$ coarse and $125$ refined), each requiring one forward-only simulation.

\smallskip
\noindent\textit{(ii) A set of candidate templates and known center.}
In this setting, the search considers the four templates used to construct the corresponding topology subset and therefore includes the ground-truth template for each evaluated object.
The coarse stage jointly evaluates the template and orientation by combining all candidate templates with 125 orientations sampled on the sphere using Fibonacci sphere sampling.
The fine stage fixes the template selected in the coarse stage and refines the orientation using 125 cone directions around the selected coarse direction.

\smallskip
\noindent\textit{(iii) Known template and center.}
In this setting, we provided the ground-truth template used to construct the evaluated object and its cavity center, leaving orientation as the remaining search variable.
We used the same coarse-to-fine orientation search as in setting (ii): 125 Fibonacci-sphere samples in the coarse stage, followed by 125 cone-direction samples around the selected coarse direction.

\smallskip
\noindent\textbf{Alternative global optimizers.}
We also evaluated alternative gradient-free optimizers for the global stage to investigate optimizer sensitivity (Section~\ref{subsec:robustness} and Appendix~\ref{subsec:global_optimizer_ablation}).
For random search in the position-search setting, the coarse grid enumeration is replaced with uniform random sampling over the same search ranges, while the fine search follows the default grid refinement around the best coarse candidate.
For Covariance Matrix Adaptation Evolution Strategy (CMA-ES)~\cite{NHansenEC2001}, the optimizer proposes hypotheses within the same parameter ranges and uses the specified evaluation budget without a separate coarse/fine split.
In the main-paper optimizer-sensitivity evaluation, all optimizer variants are followed by the same Lagrangian particle refinement stage.

\smallskip
\noindent\textbf{Relaxed-assumption settings.}
In relaxed-assumption settings (Section~\ref{subsec:robustness} and Appendix~\ref{subsec:relaxing_assumptions}), we expanded the hypothesis space with additional continuous variables, such as cavity size or physical parameters, to evaluate robustness under relaxed assumptions.
Table~\ref{tab:search_ranges} summarizes their search ranges and grid resolutions.

\begin{table}[htbp]
  \centering
  \small
  \setlength{\tabcolsep}{4.5pt}
  \begin{tabular}{@{}lccc@{}}
    \toprule
    Parameter & Range & Coarse & Fine \\
    \midrule
    Cavity size $s_c$     & $[0,\,0.8]$ & 5 & 5 \\
    Young's modulus $E$   & $[5{\times}10^5,\,5{\times}10^6]$ & 5 & 5 \\
    Poisson's ratio $\nu$ & $[0.1,\,0.4]$ & 5 & 5 \\
    \bottomrule
  \end{tabular}
  \vspace{-2mm}
  \caption{Search ranges and grid resolutions for additional continuous variables in relaxed-assumption experiments.
    Each parameter is explored using coarse-to-fine grid enumeration with the number of candidates shown for each stage.
    The candidate sets exclude the exact ground-truth values to avoid evaluating a trivially correct hypothesis.}
  \label{tab:search_ranges}
\end{table}

\subsubsection{Step (3). Lagrangian Particle Refinement}
\label{subsubsec:lagrangian_particle_refinement}

After global hypothesis optimization selected a cavity hypothesis, we refined the corresponding particle configuration following Lagrangian particle optimization~\cite{TKanekoCVPR2024}.
Starting from $\hat{\mathcal{P}}_{h^\ast}(t_1)$, the local stage treats the particles as a Lagrangian system and optimizes their initial positions through differentiable MPM simulation~\cite{YHuICLR2020}.
Observation-loss gradients are backpropagated through the full simulated sequence, allowing particle-level refinement under the physical constraints of the simulator.

Unless otherwise stated, the objective follows Eq.~\ref{eq:local_total_loss} in the main paper and combines the 3D surface loss, 2D mask loss, repulsive loss, and volume-preservation loss.
For the repulsive loss, neighboring particle pairs in $\mathcal{L}_{\mathrm{rep}}$ are defined within a radius of $1.5d_{\mathrm{grid}}$ in the initial configuration.
For the volume-preservation loss, $\mathcal{V}(\cdot)$ in $\mathcal{L}_{\mathrm{vol}}$ denotes the material volume estimated by Delaunay tetrahedralization, with $\epsilon=10^{-6}$.
We used $\lambda_{\mathrm{2D}} = 1$, $\lambda_{\mathrm{rep}} = 1$, and $\lambda_{\mathrm{vol}} = 10^4$.
For the adaptive volume-preservation loss, the reference configuration $\hat{\mathcal{P}}_{\mathrm{ref}}(t_1)$ is updated every five iterations.
We performed optimization for $2{,}000$ iterations using Adam~\cite{DPKingmaICLR2015}, with learning rate $0.0003125$ ($= d_{\mathrm{grid}} / 32$) and momentum parameters $\beta_1 = 0.9$ and $\beta_2 = 0.999$.

\subsubsection{Re-optimizing Gaussian Attributes}
\label{subsubsec:relearning}

The particle positions obtained by our method define the optimized geometry but do not directly provide the Gaussian appearance attributes required for novel-view rendering.
We therefore ran a fixed-point-cloud 3DGS optimization following the official GIC implementation~\cite{JCaiNeurIPS2024}: the particle positions were kept fixed, and only the Gaussian appearance-related attributes were optimized.
This strategy enables novel-view rendering of the optimized and predicted states.

\section{Additional Experiments and Analysis}
\label{sec:additional_experiments}

This appendix provides additional experiments and analysis to further characterize the proposed hypothesis-space reasoning framework.
We first analyze global optimizers and Lagrangian particle refinement in more detail (Appendix~\ref{subsec:global_optimizer_ablation}).
We then provide supplementary results for the interior-topology stress tests, including F-score results and an additional ablation on the asymmetric subset (Appendix~\ref{subsec:additional_topology_results}).
We also examine the effect of initialization source to separate the contribution of the optimization framework from that of the initialization representation (Appendix~\ref{subsec:init_source}).
Additional future-prediction results on the SfC and Topology datasets are provided in Appendix~\ref{subsec:additional_future_prediction}.
Finally, we evaluate robustness under relaxed structural and physical assumptions (Appendix~\ref{subsec:relaxing_assumptions}), analyze sensitivity to physical-parameter errors and material variations (Appendices~\ref{subsec:physical_parameter_errors} and~\ref{subsec:material_variation}), and report runtime details (Appendix~\ref{subsec:runtime}).

\subsection{Global Optimizer and Refinement Ablation}
\label{subsec:global_optimizer_ablation}

\begin{table*}[htbp]
  \centering  
  \setlength{\tabcolsep}{4.5pt}
  \small
  \begin{tabular}{@{}lcccccc@{}}
    \toprule
    Configuration
    & Optimizer & Global evals & Fine stage & Local refine
    & CD$\downarrow$ & F$\uparrow$ \\
    \midrule
    Instantiation only
    & -- & 0 & -- & No
    & 0.204 & 0.136 \\
    +Local refine
    & -- & 0 & -- & Yes
    & 0.156 & 0.138 \\
    \midrule
    +Global opt. (coarse)
    & Grid & 125 & No & No
    & 0.155 & 0.138 \\
    +Global opt. (coarse) + Local refine
    & Grid & 125 & No & Yes
    & 0.135 & 0.139 \\
    +Global opt. (coarse+fine)
    & Grid & 250 & Grid & No
    & 0.121 & 0.141 \\
    +Global opt. (coarse+fine) + Local refine
    & Grid & 250 & Grid & Yes
    & \textbf{0.109} & \textbf{0.144} \\
    \midrule
    +Random coarse
    & Random & 125 & No & No
    & 0.152 & 0.139 \\
    +Random coarse + Local refine
    & Random & 125 & No & Yes
    & 0.133 & 0.140 \\
    +Random coarse + grid fine
    & Random/Grid & 250 & Grid & No
    & 0.129 & 0.141 \\
    +Random coarse + grid fine + Local refine
    & Random/Grid & 250 & Grid & Yes
    & 0.112 & 0.142 \\
    \midrule
    +CMA-ES
    & CMA-ES & 125 & No & No
    & 0.124 & 0.141 \\
    +CMA-ES + Local refine
    & CMA-ES & 125 & No & Yes
    & \textbf{0.109} & 0.143 \\
    +CMA-ES
    & CMA-ES & 250 & No & No
    & 0.125 & 0.141 \\
    +CMA-ES + Local refine
    & CMA-ES & 250 & No & Yes
    & 0.110 & 0.142 \\
    \bottomrule
  \end{tabular}
  \vspace{-2mm}
  \caption{Ablation study of global optimizers and Lagrangian particle refinement on the SfC benchmark under cavity position variation.
    \emph{Global evals} denotes the number of forward-only MPM simulations used in the global stage.
    \emph{Fine stage} indicates whether grid refinement is applied after the initial global search.
    \emph{Local refine} indicates whether the selected hypothesis is refined with differentiable MPM simulation.
    All global optimizers use the same objective and search ranges.}
  \label{tab:global_optimizer_ablation}
  \vspace{-2mm}
\end{table*}

We analyzed the contributions of global hypothesis optimization and local refinement under cavity position variation, where cavity localization is most ambiguous.
This analysis extends the framework-stage ablation in Section~\ref{subsec:ablation} and the optimizer-sensitivity evaluation in Section~\ref{subsec:robustness} of the main paper.

We compared the default coarse-to-fine grid enumeration with random search and CMA-ES~\cite{NHansenEC2001}.
For grid and random search, the 125-evaluation setting uses only the coarse stage, whereas the 250-evaluation setting uses 125 coarse evaluations followed by 125 fine evaluations around the best coarse candidate.
Random search replaces coarse grid enumeration with uniform sampling over the same search ranges.
CMA-ES uses the same evaluation budgets without a separate coarse/fine split.
All global optimizer configurations use the same objective and search ranges.
For each configuration, we evaluated the selected hypothesis with and without Lagrangian particle refinement.

Table~\ref{tab:global_optimizer_ablation} shows three main trends:

\smallskip
\noindent\textit{(i) Limitations of local refinement.}
Local refinement alone improves over instantiation only, but remains limited by the initial cavity placement.

\smallskip
\noindent\textit{(ii) Effects of global hypothesis optimization.}
Adding global hypothesis optimization improves reconstruction before refinement, and combining global optimization with Lagrangian particle refinement gives the strongest results.
For the default grid optimizer, the coarse-to-fine search improves over the coarse-only search, showing that the fine search stage is useful in this hypothesis space.

\smallskip
\noindent\textit{(iii) Optimizer sensitivity.}
Alternative gradient-free optimizers achieve comparable performance under matched or comparable evaluation budgets.
CMA-ES reaches a strong solution with only 125 evaluations, but increasing its budget to 250 evaluations does not further improve performance in this low-dimensional hypothesis space.
The default coarse-to-fine grid enumeration achieves comparable final accuracy, indicating that it is a reasonable and reproducible choice for the current hypothesis space.
Overall, these results suggest that the performance gains primarily arise from hypothesis-level evaluation and subsequent Lagrangian particle refinement, rather than from the choice of global optimizer.

\subsection{Additional Interior-Topology Results}
\label{subsec:additional_topology_results}

We provide supplementary results for the interior-topology stress tests in Section~\ref{subsec:topology_stress} of the main paper.
We first report F-score results corresponding to the main-paper CD results (Appendix~\ref{subsubsec:topology_fscore}), and then provide an additional ablation on the asymmetric subset to analyze how the global and local stages behave when the dominant ambiguity lies in cavity shape (Appendix~\ref{subsubsec:topology_ablation}).

\subsubsection{F-score Results for Interior-Topology Stress Tests}
\label{subsubsec:topology_fscore}

Table~\ref{tab:topology_fscore} reports F-score results for the interior-topology stress tests corresponding to the CD results in Section~\ref{subsec:topology_stress}.
Overall, the F-score results show a trend consistent with the CD results in the main paper:
Under the known-size setting, our method improves over SfC-NeRF across all topology subsets.
When stronger structural assumptions are available, the F-score remains stable or improves across the topology subsets.

Compared with CD, the difference between the template-candidate setting and the known-template setting is smaller in F-score.
This is consistent with F-score being a thresholded metric that is less sensitive than CD to fine geometric deviations.
Although this thresholding compresses the gap between the two stronger-assumption settings, the overall trend remains consistent with the main-paper CD results.
Thus, the F-score results provide complementary support for the same overall trend.

\begin{table}[htbp]
  \centering
  \setlength{\tabcolsep}{4.5pt}
  \small
  \begin{tabular}{llccc}
    \toprule
    Method & Setting & Sym & Asym & Comp \\
    \midrule
    SfC-NeRF~\cite{TKanekoCVPR2025}
           & Size
                     & 0.085 & 0.084 & 0.081 \\
    Ours
           & Size
                     & \textbf{0.089} & \textbf{0.090} & \textbf{0.085} \\
    \midrule
    Ours
           & Cand. w/ center
                     & 0.091 & 0.096 & 0.095 \\
    Ours
           & Template w/ center
                     & 0.092 & 0.096 & 0.095 \\
    \bottomrule
  \end{tabular}
  \vspace{-2mm}
  \caption{F-score results for the interior-topology stress tests.
    Results correspond to the CD results reported in Section~\ref{subsec:topology_stress} of the main paper.
    Higher values indicate better thresholded reconstruction consistency on the symmetric (Sym), asymmetric (Asym), and complex (Comp) subsets.
    Bold indicates the better result between SfC-NeRF and Ours under the known-size setting.}
  \label{tab:topology_fscore}
  \vspace{-2mm}
\end{table}

\subsubsection{Ablation on the Asymmetric Topology Subset}
\label{subsubsec:topology_ablation}

We further analyzed the asymmetric subset of the Topology dataset to complement the interior-topology stress tests in Section~\ref{subsec:topology_stress} of the main paper.
We chose this subset because its cavities break symmetry and introduce preferred orientations, creating a clear shape mismatch with the inner-surface initialization derived from the external geometry.
Unlike the cavity-position setting in the SfC benchmark, this setting primarily changes the cavity shape rather than its location.
This makes the experiment useful for examining how the global and local stages behave when the dominant ambiguity lies in cavity shape.

Table~\ref{tab:topology_ablation} provides an ablation complementary to the position-variation analysis in the main paper.

\smallskip
\noindent\textbf{Stage-wise ablation.}
Local refinement alone gives a large improvement over instantiation only, suggesting that particle-level adjustment is particularly useful when the initial hypothesis has a shape mismatch.
Global optimization alone also improves over instantiation only, although its effect is smaller than local refinement in this shape-mismatch setting.
The full pipeline achieves the best CD, indicating that global hypothesis optimization and Lagrangian particle refinement remain complementary, while their relative contributions depend on the dominant source of ambiguity.

\smallskip
\noindent\textbf{Geometry-aware losses.}
The fixed volume-preservation loss slightly degrades CD, likely because asymmetric cavities require larger shape adjustment and a fixed reference volume can be overly restrictive.
The adaptive volume-preservation loss improves reconstruction by updating the reference configuration during refinement.
F-score changes only marginally across configurations, suggesting that it is less sensitive than CD to the geometric differences observed in this setting.

\begin{table}[htbp]
  \centering  
  \setlength{\tabcolsep}{4.5pt}
  \small
  \begin{tabular}{lcc}
    \toprule
    Configuration & CD$\downarrow$ & F$\uparrow$ \\
    \midrule
    \multicolumn{3}{l}{\emph{(a) Framework stages}} \\
    Instantiation only                    & 0.359 & 0.086 \\
    +Local refine                         & 0.221 & \textbf{0.090} \\
    +Global opt.                          & 0.273 & 0.088 \\
    +Global opt. + Local refine (Full)    & \textbf{0.185} & \textbf{0.090} \\
    \midrule
    \multicolumn{3}{l}{\emph{(b) Geometry-aware losses}} \\
    None                                  & 0.240 & 0.085 \\
    Rep. only                             & 0.213 & \textbf{0.090} \\
    Rep. + Fixed volume                   & 0.243 & \textbf{0.090} \\
    Rep. + Adaptive volume (Full)         & \textbf{0.185} & \textbf{0.090} \\
    \bottomrule
  \end{tabular}
  \vspace{-2mm}
  \caption{Additional ablation results on the asymmetric subset of the Topology dataset.
    (a) Effect of framework stages.
    (b) Effect of geometry-aware losses during Lagrangian particle refinement.
    CD values are scaled by $10^3$.}
  \label{tab:topology_ablation}
\end{table}

\subsection{Effect of Initialization Source}
\label{subsec:init_source}

SfC-NeRF uses Direct Voxel Grid Optimization (DVGO)~\cite{CSunCVPR2022}, a voxel-based NeRF representation, for initialization, whereas our default implementation uses 3D Gaussian Splatting (3DGS)~\cite{BKerblTOG2023}.
To isolate the influence of the initialization source, we additionally initialized our optimization from DVGO reconstructions by thresholding the learned alpha grid and sampling particle positions from occupied voxels.
For this DVGO-based initialization variant, the deformation network used for surface extraction is unavailable, so we disabled the 3D surface loss.

Table~\ref{tab:init_source} compares different combinations of initialization sources and optimization frameworks.
Even when initialized from the same DVGO reconstruction used by SfC-NeRF, our optimization framework improves over SfC-NeRF.
Using the default 3DGS initialization further improves reconstruction accuracy.
These results indicate that the proposed optimization framework contributes substantially to the performance gain, while the initialization source provides additional improvement.

\begin{table}[htbp]
  \centering
  \setlength{\tabcolsep}{4.5pt}
  \small
  \begin{tabular}{llcccc}
    \toprule
    & & \multicolumn{2}{c}{Size} & \multicolumn{2}{c}{Position} \\
    \cmidrule(lr){3-4}\cmidrule(lr){5-6}
    Initialization & Optimization
      & CD$\downarrow$ & F$\uparrow$
                   & CD$\downarrow$ & F$\uparrow$ \\
    \midrule
    DVGO & SfC-NeRF
      & 0.165 & 0.135 & 0.281 & 0.128 \\
    DVGO & Ours
      & 0.104 & 0.143 & 0.170 & 0.136 \\
    3DGS & Ours
      & \textbf{0.063} & \textbf{0.151} & \textbf{0.109} & \textbf{0.144} \\
    \bottomrule
  \end{tabular}
  \vspace{-2mm}
  \caption{Reconstruction accuracy under different initialization sources and optimization frameworks.
    Even with the same DVGO initialization as SfC-NeRF, our hypothesis-space reasoning framework improves accuracy.
    CD values are scaled by $10^3$.}
  \label{tab:init_source}
  \vspace{-2mm}
\end{table}

\subsection{Additional Future Prediction Results}
\label{subsec:additional_future_prediction}

\begin{table*}[htbp]
  \centering
  \setlength{\tabcolsep}{4.5pt}
  \small
  \begin{tabular}{@{}lcccccccc@{}}
    \toprule
    & \multicolumn{4}{c}{Size}
    & \multicolumn{4}{c}{Position} \\
    \cmidrule(lr){2-5}
    \cmidrule(lr){6-9}
    Method
    & \multicolumn{2}{c}{$+7$}
    & \multicolumn{2}{c}{$+14$}
    & \multicolumn{2}{c}{$+7$}
    & \multicolumn{2}{c}{$+14$} \\
    \cmidrule(lr){2-3}\cmidrule(lr){4-5}
    \cmidrule(lr){6-7}\cmidrule(lr){8-9}
    & PSNR$\uparrow$ & SSIM$\uparrow$
    & PSNR$\uparrow$ & SSIM$\uparrow$
    & PSNR$\uparrow$ & SSIM$\uparrow$
    & PSNR$\uparrow$ & SSIM$\uparrow$ \\
    \midrule
    SfC-NeRF~\cite{TKanekoCVPR2025}
    & 30.8 & 0.987
    & 29.0 & 0.984
    & 27.1 & 0.982
    & 25.2 & 0.978 \\
    GIC~\cite{JCaiNeurIPS2024}
    & 28.4 & 0.985
    & 26.7 & 0.982
    & 23.1 & 0.977
    & 22.1 & 0.974 \\
    Ours (Global only)
    & 29.4 & 0.986
    & 27.3 & 0.982
    & 24.9 & 0.980
    & 23.6 & 0.977 \\
    Ours (Full)
    & \textbf{37.0} & \textbf{0.995}
    & \textbf{35.5} & \textbf{0.993}
    & \textbf{33.6} & \textbf{0.991}
    & \textbf{30.8} & \textbf{0.988} \\
    \bottomrule
  \end{tabular}
  \vspace{-2mm}
  \caption{Future prediction accuracy on the SfC benchmark under cavity size and position variations.
    All methods are optimized using 14 observed frames and evaluated at $+7$ and $+14$ future frames.}
  \label{tab:future_sfc}
\end{table*}

\begin{table*}[htbp]
  \centering
  \setlength{\tabcolsep}{4.5pt}
  \small
  \begin{tabular}{@{}lcccccccccccc@{}}
    \toprule
    & \multicolumn{4}{c}{Symmetric}
    & \multicolumn{4}{c}{Asymmetric}
    & \multicolumn{4}{c}{Complex} \\
    \cmidrule(lr){2-5}
    \cmidrule(lr){6-9}
    \cmidrule(lr){10-13}
    Method
    & \multicolumn{2}{c}{$+7$}
    & \multicolumn{2}{c}{$+14$}
    & \multicolumn{2}{c}{$+7$}
    & \multicolumn{2}{c}{$+14$}
    & \multicolumn{2}{c}{$+7$}
    & \multicolumn{2}{c}{$+14$} \\
    \cmidrule(lr){2-3}\cmidrule(lr){4-5}
    \cmidrule(lr){6-7}\cmidrule(lr){8-9}
    \cmidrule(lr){10-11}\cmidrule(lr){12-13}
    & PSNR$\uparrow$ & SSIM$\uparrow$
    & PSNR$\uparrow$ & SSIM$\uparrow$
    & PSNR$\uparrow$ & SSIM$\uparrow$
    & PSNR$\uparrow$ & SSIM$\uparrow$
    & PSNR$\uparrow$ & SSIM$\uparrow$
                     & PSNR$\uparrow$ & SSIM$\uparrow$ \\
    \midrule
    SfC-NeRF~\cite{TKanekoCVPR2025}
    & 31.4 & 0.988
    & 30.0 & 0.987
    & 29.1 & 0.986
    & 27.5 & 0.984
    & 28.8 & 0.985
                     & 26.7 & 0.982 \\
    GIC~\cite{JCaiNeurIPS2024}
    & 24.3 & 0.981
    & 23.9 & 0.980
    & 22.1 & 0.977
    & 21.5 & 0.975
    & 21.4 & 0.975
                     & 20.5 & 0.973 \\
    Ours (Global only)
    & 25.5 & 0.982
    & 24.9 & 0.981
    & 23.4 & 0.979
    & 22.4 & 0.976
    & 22.1 & 0.976
                     & 21.0 & 0.973 \\
    Ours (Full)
    & \textbf{35.7} & \textbf{0.993}
    & \textbf{33.0} & \textbf{0.989}
    & \textbf{34.5} & \textbf{0.992}
    & \textbf{30.8} & \textbf{0.987}
    & \textbf{34.7} & \textbf{0.992}
                     & \textbf{31.1} & \textbf{0.988} \\
    \bottomrule
  \end{tabular}
  \vspace{-2mm}
  \caption{Future prediction accuracy on the Topology dataset under symmetric, asymmetric, and complex subsets.
    All methods are optimized using 14 observed frames and evaluated at $+7$ and $+14$ future frames.}
  \label{tab:future_topology}
  \vspace{-3mm}
\end{table*}

We report additional quantitative results for future prediction on the SfC and Topology datasets.
These results complement the analysis in Section~\ref{subsec:future_prediction} of the main paper.
All methods are optimized using $N = 14$ observed frames and evaluated at $+7$ and $+14$ future frames.
GIC and our variants use the GIC-based rendering pipeline described in Appendix~\ref{subsubsec:relearning}, whereas SfC-NeRF uses its original PAC-NeRF-based rendering pipeline~\cite{XLiICLR2023}.

\smallskip
\noindent\textit{SfC benchmark.}
On the SfC benchmark, Table~\ref{tab:future_sfc} reports results under cavity size and position variations.
Our full model achieves the best performance for all settings and prediction horizons.
The gains are substantial under both variations, demonstrating the strong end-to-end future-prediction performance of our full framework.
For example, at $+14$ frames under position variation, our method improves PSNR by more than 5 dB over SfC-NeRF.

\smallskip
\noindent\textit{Topology dataset.}
Table~\ref{tab:future_topology} reports results on the Topology dataset under symmetric, asymmetric, and complex subsets.
Our full model outperforms all baselines for every subset and prediction horizon.
The gap between the global-only variant and the full model indicates that Lagrangian particle refinement is important for future prediction, especially for complex interior structures.

\smallskip
Together, these results show that global hypothesis optimization and Lagrangian particle refinement provide complementary benefits for future prediction under diverse geometry and topology conditions.

\subsection{Robustness under Relaxed Assumptions}
\label{subsec:relaxing_assumptions}

We evaluated robustness by relaxing controlled assumptions used in the main experiments.
Specifically, we considered two settings: \textit{unknown cavity size} and \textit{unknown physical parameters}.
For each setting, global hypothesis optimization is extended to include the corresponding variables.
The search ranges and grid resolutions are summarized in Table~\ref{tab:search_ranges} in Appendix~\ref{sec:implementation}.
These relaxed settings test whether the framework remains effective when part of the structural or physical information is not provided.

\subsubsection{Unknown Cavity Size}
\label{subsubsec:unknown_cavity_size}

We evaluated the unknown-size setting on the SfC benchmark under both cavity size and position variations.
The cavity size is estimated jointly with cavity position by extending the global hypothesis space to include size as an additional variable.
For SfC-NeRF, the mass loss term is removed to avoid relying on the known-size assumption.
Table~\ref{tab:relax_size} summarizes the results.
Compared with the known-size setting, the unknown-size setting degrades reconstruction accuracy, but our method remains substantially better than SfC-NeRF under both assumption settings.
These results indicate that the proposed framework can incorporate cavity size into hypothesis-level optimization without requiring it to be fixed in advance.

\begin{table}[htbp]
  \centering
  \setlength{\tabcolsep}{4.5pt}
  \small
  \begin{tabular}{llcccc}
    \toprule
    & & \multicolumn{2}{c}{Size}
    & \multicolumn{2}{c}{Position} \\
    \cmidrule(lr){3-4}\cmidrule(lr){5-6}
    Method & Size
      & CD$\downarrow$ & F$\uparrow$
           & CD$\downarrow$ & F$\uparrow$ \\
    \midrule
    SfC-NeRF & Known
      & 0.165 & 0.135
           & 0.281 & 0.128 \\
    Ours & Known
      & \textbf{0.063} & \textbf{0.151}
           & \textbf{0.109} & \textbf{0.144} \\
    \midrule
    SfC-NeRF & Unknown
      & 0.637 & 0.125
           & 0.606 & 0.121 \\
    Ours & Unknown
      & \textbf{0.086} & \textbf{0.147}
           & \textbf{0.128} & \textbf{0.141} \\
    \bottomrule
  \end{tabular}
  \vspace{-2mm}
  \caption{Results when relaxing the known-size assumption.
    \emph{Known} indicates that the cavity size is provided, while \emph{Unknown} estimates the cavity size jointly with cavity position.
    Bold indicates the better result between SfC-NeRF and Ours under each assumption setting.
    CD values are scaled by $10^3$.}
  \label{tab:relax_size}
\end{table}

\subsubsection{Unknown Physical Parameters}
\label{subsubsec:unknown_physical_parameters}

We evaluated robustness when the physical parameters, namely Young's modulus and Poisson's ratio, are unknown.
The global hypothesis space is extended to jointly search cavity position and physical parameters, both of which are further optimized during local refinement.
For SfC-NeRF, the physical parameters are optimized together with the reconstruction variables, using the ground-truth values as initialization to provide a favorable condition.
Table~\ref{tab:relax_phys} shows that our method exhibits only minor degradation compared with the known-parameter setting while outperforming SfC-NeRF under both assumption settings.
Table~\ref{tab:relax_phys_param_error} further shows that our method achieves lower estimation errors for both Young's modulus and Poisson's ratio.

\begin{table}[htbp]
  \centering
  \setlength{\tabcolsep}{4.5pt}
  \small
  \begin{tabular}{llcccc}
    \toprule
    & & \multicolumn{2}{c}{Size}
    & \multicolumn{2}{c}{Position} \\
    \cmidrule(lr){3-4}\cmidrule(lr){5-6}
    Method & Phys. Params.
      & CD$\downarrow$ & F$\uparrow$
           & CD$\downarrow$ & F$\uparrow$ \\
    \midrule
    SfC-NeRF & Known
      & 0.165 & 0.135
           & 0.281 & 0.128 \\
    Ours & Known
      & \textbf{0.063} & \textbf{0.151}
           & \textbf{0.109} & \textbf{0.144} \\
    \midrule
    SfC-NeRF & Unknown
      & 0.181 & 0.135
           & 0.321 & 0.126 \\
    Ours & Unknown
      & \textbf{0.063} & \textbf{0.151}
           & \textbf{0.121} & \textbf{0.142} \\
    \bottomrule
  \end{tabular}
  \vspace{-2mm}
  \caption{Results when relaxing the physical-parameter assumption.
    \emph{Known} indicates that the physical parameters are provided, while \emph{Unknown} estimates them jointly with cavity position.
    Bold indicates the better result between SfC-NeRF and Ours under each assumption setting.
    CD values are scaled by $10^3$.}
  \label{tab:relax_phys}
  \vspace{-2mm}
\end{table}

\begin{table}[htbp]
  \centering
  \setlength{\tabcolsep}{4.5pt}
  \small
  \begin{tabular}{lcccc}
    \toprule
    & \multicolumn{2}{c}{Size}
    & \multicolumn{2}{c}{Position} \\
    \cmidrule(lr){2-3}\cmidrule(lr){4-5}
    Method
    & $\log_{10}E\downarrow$ & $\nu\downarrow$
    & $\log_{10}E\downarrow$ & $\nu\downarrow$ \\
    \midrule
    SfC-NeRF
    & 0.052 & 0.201
    & 0.106 & 0.180 \\
    Ours
    & \textbf{0.023} & \textbf{0.119}
    & \textbf{0.068} & \textbf{0.162} \\
    \bottomrule
  \end{tabular}
  \vspace{-2mm}
  \caption{Physical-parameter estimation errors in the unknown-parameter setting.
    We report the mean absolute errors of Young's modulus $\log_{10}E$ and Poisson's ratio $\nu$ under the size- and position-variation settings of the SfC benchmark.}
  \label{tab:relax_phys_param_error}
\end{table}

\noindent\textbf{Summary.}
The proposed framework remains stable when cavity size or physical parameters are not provided.
These results support the robustness of hypothesis-space reasoning under relaxed structural and physical assumptions.

\subsection{Robustness to Physical-Parameter Errors}
\label{subsec:physical_parameter_errors}

We further evaluated sensitivity to physical-parameter misspecification, extending the representative $\pm10\%$ error results reported in Table~\ref{tab:robustness} of the main paper.
Unlike the unknown-parameter setting in Appendix~\ref{subsec:relaxing_assumptions}, the physical parameters are provided to the method but are perturbed from the ground-truth values.
We evaluated larger perturbations of $\pm20\%$ and $\pm30\%$ for Young's modulus $E$ and Poisson's ratio $\nu$ on sphere objects under cavity size and position variations.

Table~\ref{tab:physical_parameter_errors} summarizes the results.
Errors in Young's modulus lead to a gradual degradation as the perturbation magnitude increases, especially under cavity position variation.
This is consistent with stiffness directly affecting post-collision deformation, so an incorrect stiffness value can make cavity localization from motion cues more difficult.
In contrast, errors in Poisson's ratio have little effect in this setting: even under $\pm30\%$ perturbations, the results remain close to the clean setting for both size and position variations.
Overall, these results indicate that the framework is robust to moderate physical-parameter errors, while large stiffness errors can degrade reconstruction when cavity localization is ambiguous.

\begin{table}[htbp]
  \centering
  \setlength{\tabcolsep}{4.5pt}
  \small
  \begin{tabular}{lcccc}
    \toprule
    & \multicolumn{2}{c}{Size}
    & \multicolumn{2}{c}{Position} \\
    \cmidrule(lr){2-3}\cmidrule(lr){4-5}
    Condition
    & CD$\downarrow$ & F$\uparrow$
    & CD$\downarrow$ & F$\uparrow$ \\
    \midrule
    Ours (Clean)
    & 0.059 & 0.097 & 0.100 & 0.093 \\
    Ours ($E$, $\pm10\%$ error)
    & 0.061 & 0.096 & 0.100 & 0.093 \\
    Ours ($E$, $\pm20\%$ error)
    & 0.069 & 0.095 & 0.120 & 0.092 \\
    Ours ($E$, $\pm30\%$ error)
    & 0.080 & 0.095 & 0.164 & 0.089 \\
    Ours ($\nu$, $\pm10\%$ error)
    & 0.059 & 0.097 & 0.101 & 0.093 \\
    Ours ($\nu$, $\pm20\%$ error)
    & 0.060 & 0.097 & 0.101 & 0.093 \\
    Ours ($\nu$, $\pm30\%$ error)
    & 0.061 & 0.097 & 0.102 & 0.093 \\
    \bottomrule
  \end{tabular}
  \vspace{-2mm}
  \caption{Sensitivity to physical-parameter errors on sphere objects under cavity size and position variations.
    Results for each error level are averaged over positive and negative perturbations.
    CD values are scaled by $10^3$.}
  \label{tab:physical_parameter_errors}
\end{table}

\subsection{Robustness to Material Variations}
\label{subsec:material_variation}

\begin{table*}[htbp]
  \vspace{9mm}
  \centering
  \setlength{\tabcolsep}{4.5pt}
  \small

  \begin{minipage}{\textwidth}
    \centering
    \begin{tabular}{lcccccccccccc}
      \toprule
      & \multicolumn{2}{c}{$2.5 \times 10^5$}
      & \multicolumn{2}{c}{$5.0 \times 10^5$}
      & \multicolumn{2}{c}{$1.0 \times 10^6$}
      & \multicolumn{2}{c}{$2.0 \times 10^6$}
      & \multicolumn{2}{c}{$4.0 \times 10^6$}
      & \multicolumn{2}{c}{Avg.} \\
      \cmidrule(lr){2-3}
      \cmidrule(lr){4-5}
      \cmidrule(lr){6-7}
      \cmidrule(lr){8-9}
      \cmidrule(lr){10-11}
      \cmidrule(lr){12-13}
      Method
      & CD$\downarrow$ & F$\uparrow$
      & CD$\downarrow$ & F$\uparrow$
      & CD$\downarrow$ & F$\uparrow$
      & CD$\downarrow$ & F$\uparrow$
      & CD$\downarrow$ & F$\uparrow$
      & CD$\downarrow$ & F$\uparrow$ \\
      \midrule
      SfC-NeRF
      & 0.206 & 0.132
      & 0.196 & 0.132
      & 0.195 & 0.133
      & 0.215 & 0.132
      & 0.230 & 0.132
      & 0.208 & 0.132 \\
      Ours
      & \textbf{0.066} & \textbf{0.150}
      & \textbf{0.062} & \textbf{0.148}
      & \textbf{0.068} & \textbf{0.150}
      & \textbf{0.069} & \textbf{0.149}
      & \textbf{0.072} & \textbf{0.147}
      & \textbf{0.068} & \textbf{0.149} \\
      \bottomrule
    \end{tabular}
    \vspace{-2mm}
    \caption*{(a) External-shape variation}
    \vspace{1mm}
  \end{minipage}

  \smallskip

  \begin{minipage}{\textwidth}
    \centering
    \begin{tabular}{lcccccccccccc}
      \toprule
      & \multicolumn{2}{c}{$2.5 \times 10^5$}
      & \multicolumn{2}{c}{$5.0 \times 10^5$}
      & \multicolumn{2}{c}{$1.0 \times 10^6$}
      & \multicolumn{2}{c}{$2.0 \times 10^6$}
      & \multicolumn{2}{c}{$4.0 \times 10^6$}
      & \multicolumn{2}{c}{Avg.} \\
      \cmidrule(lr){2-3}
      \cmidrule(lr){4-5}
      \cmidrule(lr){6-7}
      \cmidrule(lr){8-9}
      \cmidrule(lr){10-11}
      \cmidrule(lr){12-13}
      Method
      & CD$\downarrow$ & F$\uparrow$
      & CD$\downarrow$ & F$\uparrow$
      & CD$\downarrow$ & F$\uparrow$
      & CD$\downarrow$ & F$\uparrow$
      & CD$\downarrow$ & F$\uparrow$
      & CD$\downarrow$ & F$\uparrow$ \\
      \midrule
      SfC-NeRF
      & 0.067 & 0.093
      & 0.069 & 0.093
      & 0.067 & 0.093
      & 0.075 & 0.093
      & 0.099 & 0.092
      & 0.075 & 0.093 \\
      Ours
      & \textbf{0.059} & \textbf{0.098}
      & \textbf{0.058} & \textbf{0.096}
      & \textbf{0.059} & \textbf{0.097}
      & \textbf{0.059} & \textbf{0.097}
      & \textbf{0.062} & \textbf{0.096}
      & \textbf{0.059} & \textbf{0.097} \\
      \bottomrule
    \end{tabular}
    \vspace{-2mm}
    \caption*{(b) Cavity-size variation}
    \vspace{1mm}
  \end{minipage}

  \smallskip

  \begin{minipage}{\textwidth}
    \centering
    \begin{tabular}{lcccccccccccc}
      \toprule
      & \multicolumn{2}{c}{$2.5 \times 10^5$}
      & \multicolumn{2}{c}{$5.0 \times 10^5$}
      & \multicolumn{2}{c}{$1.0 \times 10^6$}
      & \multicolumn{2}{c}{$2.0 \times 10^6$}
      & \multicolumn{2}{c}{$4.0 \times 10^6$}
      & \multicolumn{2}{c}{Avg.} \\
      \cmidrule(lr){2-3}
      \cmidrule(lr){4-5}
      \cmidrule(lr){6-7}
      \cmidrule(lr){8-9}
      \cmidrule(lr){10-11}
      \cmidrule(lr){12-13}
      Method
      & CD$\downarrow$ & F$\uparrow$
      & CD$\downarrow$ & F$\uparrow$
      & CD$\downarrow$ & F$\uparrow$
      & CD$\downarrow$ & F$\uparrow$
      & CD$\downarrow$ & F$\uparrow$
      & CD$\downarrow$ & F$\uparrow$ \\
      \midrule
      SfC-NeRF
      & 0.165 & 0.086
      & 0.149 & 0.087
      & 0.152 & 0.087
      & 0.187 & 0.086
      & 0.168 & 0.086
      & 0.164 & 0.087 \\
      Ours
      & \textbf{0.086} & \textbf{0.093}
      & \textbf{0.091} & \textbf{0.093}
      & \textbf{0.100} & \textbf{0.093}
      & \textbf{0.128} & \textbf{0.091}
      & \textbf{0.119} & \textbf{0.092}
      & \textbf{0.105} & \textbf{0.092} \\
      \bottomrule
    \end{tabular}
    \vspace{-2mm}
    \caption*{(c) Cavity-position variation}
    \vspace{-2mm}
  \end{minipage}

  \caption{Robustness to Young's modulus across geometry settings.
    We report CD$\downarrow$ and F-score (F$\uparrow$) under (a) external-shape, (b) cavity-size, and (c) cavity-position variations.
    Avg. represents the average over the five Young's modulus settings.
    CD values are scaled by $10^3$.}
  \label{tab:young_material_variation}
  \vspace{9mm}
\end{table*}

\begin{table*}[htbp]
  \centering
  \setlength{\tabcolsep}{4.5pt}
  \small

  \begin{minipage}{\textwidth}
    \centering
    \begin{tabular}{lcccccccccccc}
      \toprule
      & \multicolumn{2}{c}{$0.2$}
      & \multicolumn{2}{c}{$0.25$}
      & \multicolumn{2}{c}{$0.3$}
      & \multicolumn{2}{c}{$0.35$}
      & \multicolumn{2}{c}{$0.4$}
      & \multicolumn{2}{c}{Avg.} \\
      \cmidrule(lr){2-3}
      \cmidrule(lr){4-5}
      \cmidrule(lr){6-7}
      \cmidrule(lr){8-9}
      \cmidrule(lr){10-11}
      \cmidrule(lr){12-13}
      Method
      & CD$\downarrow$ & F$\uparrow$
      & CD$\downarrow$ & F$\uparrow$
      & CD$\downarrow$ & F$\uparrow$
      & CD$\downarrow$ & F$\uparrow$
      & CD$\downarrow$ & F$\uparrow$
      & CD$\downarrow$ & F$\uparrow$ \\
      \midrule
      SfC-NeRF
      & 0.212 & 0.133
      & 0.214 & 0.131
      & 0.195 & 0.133
      & 0.225 & 0.132
      & 0.226 & 0.131
      & 0.214 & 0.132 \\
      Ours
      & \textbf{0.064} & \textbf{0.150}
      & \textbf{0.067} & \textbf{0.148}
      & \textbf{0.068} & \textbf{0.150}
      & \textbf{0.067} & \textbf{0.149}
      & \textbf{0.066} & \textbf{0.149}
      & \textbf{0.066} & \textbf{0.149} \\
      \bottomrule
    \end{tabular}
    \vspace{-2mm}
    \caption*{(a) External-shape variation}
    \vspace{1mm}
  \end{minipage}

  \smallskip

  \begin{minipage}{\textwidth}
    \centering
    \begin{tabular}{lcccccccccccc}
      \toprule
      & \multicolumn{2}{c}{$0.2$}
      & \multicolumn{2}{c}{$0.25$}
      & \multicolumn{2}{c}{$0.3$}
      & \multicolumn{2}{c}{$0.35$}
      & \multicolumn{2}{c}{$0.4$}
      & \multicolumn{2}{c}{Avg.} \\
      \cmidrule(lr){2-3}
      \cmidrule(lr){4-5}
      \cmidrule(lr){6-7}
      \cmidrule(lr){8-9}
      \cmidrule(lr){10-11}
      \cmidrule(lr){12-13}
      Method
      & CD$\downarrow$ & F$\uparrow$
      & CD$\downarrow$ & F$\uparrow$
      & CD$\downarrow$ & F$\uparrow$
      & CD$\downarrow$ & F$\uparrow$
      & CD$\downarrow$ & F$\uparrow$
      & CD$\downarrow$ & F$\uparrow$ \\
      \midrule
      SfC-NeRF
      & 0.070 & 0.093
      & 0.076 & 0.092
      & 0.067 & 0.093
      & 0.073 & 0.093
      & 0.066 & 0.092
      & 0.070 & 0.093 \\
      Ours
      & \textbf{0.059} & \textbf{0.097}
      & \textbf{0.059} & \textbf{0.096}
      & \textbf{0.059} & \textbf{0.097}
      & \textbf{0.058} & \textbf{0.097}
      & \textbf{0.058} & \textbf{0.098}
      & \textbf{0.058} & \textbf{0.097} \\
      \bottomrule
    \end{tabular}
    \vspace{-2mm}
    \caption*{(b) Cavity-size variation}
    \vspace{1mm}
  \end{minipage}

  \smallskip

  \begin{minipage}{\textwidth}
    \centering
    \begin{tabular}{lcccccccccccc}
      \toprule
      & \multicolumn{2}{c}{$0.2$}
      & \multicolumn{2}{c}{$0.25$}
      & \multicolumn{2}{c}{$0.3$}
      & \multicolumn{2}{c}{$0.35$}
      & \multicolumn{2}{c}{$0.4$}
      & \multicolumn{2}{c}{Avg.} \\
      \cmidrule(lr){2-3}
      \cmidrule(lr){4-5}
      \cmidrule(lr){6-7}
      \cmidrule(lr){8-9}
      \cmidrule(lr){10-11}
      \cmidrule(lr){12-13}
      Method
      & CD$\downarrow$ & F$\uparrow$
      & CD$\downarrow$ & F$\uparrow$
      & CD$\downarrow$ & F$\uparrow$
      & CD$\downarrow$ & F$\uparrow$
      & CD$\downarrow$ & F$\uparrow$
      & CD$\downarrow$ & F$\uparrow$ \\
      \midrule
      SfC-NeRF
      & 0.202 & 0.087
      & 0.159 & 0.086
      & 0.152 & 0.087
      & 0.199 & 0.087
      & 0.176 & 0.086
      & 0.178 & 0.087 \\
      Ours
      & \textbf{0.104} & \textbf{0.093}
      & \textbf{0.100} & \textbf{0.094}
      & \textbf{0.100} & \textbf{0.093}
      & \textbf{0.100} & \textbf{0.094}
      & \textbf{0.098} & \textbf{0.093}
      & \textbf{0.101} & \textbf{0.093} \\
      \bottomrule
    \end{tabular}
    \vspace{-2mm}
    \caption*{(c) Cavity-position variation}
    \vspace{-2mm}
  \end{minipage}

  \caption{Robustness to Poisson's ratio across geometry settings.
    We report CD$\downarrow$ and F-score (F$\uparrow$) under (a) external-shape, (b) cavity-size, and (c) cavity-position variations.
    Avg. represents the average over the five Poisson's ratio settings.
    CD values are scaled by $10^3$.}
  \label{tab:poisson_material_variation}
  \vspace{9mm}
\end{table*}

\begin{table*}[htbp]
  \centering
  \setlength{\tabcolsep}{4.5pt}
  \small

  \begin{minipage}{\textwidth}
    \centering
    \begin{tabular}{lcccccccccc}
      \toprule
      & \multicolumn{2}{c}{Newtonian}
      & \multicolumn{2}{c}{Non-Newtonian}
      & \multicolumn{2}{c}{Plasticine}
      & \multicolumn{2}{c}{Sand}
      & \multicolumn{2}{c}{Avg.} \\
      \cmidrule(lr){2-3}
      \cmidrule(lr){4-5}
      \cmidrule(lr){6-7}
      \cmidrule(lr){8-9}
      \cmidrule(lr){10-11}
      Method
      & CD$\downarrow$ & F$\uparrow$
      & CD$\downarrow$ & F$\uparrow$
      & CD$\downarrow$ & F$\uparrow$
                       & CD$\downarrow$ & F$\uparrow$
                       & CD$\downarrow$ & F$\uparrow$ \\
      \midrule
      SfC-NeRF
      & 0.196 & 0.131
      & 0.218 & 0.132
      & 0.230 & 0.132
                       & 0.222 & 0.132
                       & 0.216 & 0.132 \\
      Ours
      & \textbf{0.076} & \textbf{0.147}
      & \textbf{0.083} & \textbf{0.146}
      & \textbf{0.086} & \textbf{0.146}
                       & \textbf{0.071} & \textbf{0.150}
                       & \textbf{0.080} & \textbf{0.147} \\
      \bottomrule
    \end{tabular}
    \vspace{-2mm}
    \caption*{(a) External-shape variation}
    \vspace{1mm}
  \end{minipage}

  \smallskip

  \begin{minipage}{\textwidth}
    \centering
    \begin{tabular}{lcccccccccc}
      \toprule
      & \multicolumn{2}{c}{Newtonian}
      & \multicolumn{2}{c}{Non-Newtonian}
      & \multicolumn{2}{c}{Plasticine}
      & \multicolumn{2}{c}{Sand}
      & \multicolumn{2}{c}{Avg.} \\
      \cmidrule(lr){2-3}
      \cmidrule(lr){4-5}
      \cmidrule(lr){6-7}
      \cmidrule(lr){8-9}
      \cmidrule(lr){10-11}
      Method
      & CD$\downarrow$ & F$\uparrow$
      & CD$\downarrow$ & F$\uparrow$
      & CD$\downarrow$ & F$\uparrow$
                       & CD$\downarrow$ & F$\uparrow$
                       & CD$\downarrow$ & F$\uparrow$ \\
      \midrule
      SfC-NeRF
      & 0.071 & 0.093
      & 0.088 & 0.092
      & 0.080 & 0.093
                       & 0.082 & 0.091
                       & 0.080 & 0.092 \\
      Ours
      & \textbf{0.063} & \textbf{0.099}
      & \textbf{0.061} & \textbf{0.098}
      & \textbf{0.079} & \textbf{0.096}
                       & \textbf{0.065} & \textbf{0.097}
                       & \textbf{0.067} & \textbf{0.097} \\
      \bottomrule
    \end{tabular}
    \vspace{-2mm}
    \caption*{(b) Cavity-size variation}
    \vspace{1mm}
  \end{minipage}

  \smallskip

  \begin{minipage}{\textwidth}
    \centering
    \begin{tabular}{lcccccccccc}
      \toprule
      & \multicolumn{2}{c}{Newtonian}
      & \multicolumn{2}{c}{Non-Newtonian}
      & \multicolumn{2}{c}{Plasticine}
      & \multicolumn{2}{c}{Sand}
      & \multicolumn{2}{c}{Avg.} \\
      \cmidrule(lr){2-3}
      \cmidrule(lr){4-5}
      \cmidrule(lr){6-7}
      \cmidrule(lr){8-9}
      \cmidrule(lr){10-11}
      Method
      & CD$\downarrow$ & F$\uparrow$
      & CD$\downarrow$ & F$\uparrow$
      & CD$\downarrow$ & F$\uparrow$
                       & CD$\downarrow$ & F$\uparrow$
                       & CD$\downarrow$ & F$\uparrow$ \\
      \midrule
      SfC-NeRF
      & 0.235 & 0.084
      & 0.216 & 0.083
      & 0.205 & 0.085
                       & 0.217 & 0.084
                       & 0.218 & 0.084 \\
      Ours
      & \textbf{0.108} & \textbf{0.094}
      & \textbf{0.095} & \textbf{0.094}
      & \textbf{0.129} & \textbf{0.092}
                       & \textbf{0.107} & \textbf{0.093}
                       & \textbf{0.110} & \textbf{0.093} \\
      \bottomrule
    \end{tabular}
    \vspace{-2mm}
    \caption*{(c) Cavity-position variation}
    \vspace{-2mm}
  \end{minipage}

  \caption{Robustness across material models and geometry settings.
    We report CD$\downarrow$ and F-score (F$\uparrow$) under (a) external-shape, (b) cavity-size, and (c) cavity-position variations.
    Results for Newtonian, non-Newtonian, and plasticine are averaged over two parameter settings per material category.
    Avg. is computed over all evaluated material settings.
    CD values are scaled by $10^3$.}
  \label{tab:material_model_variation}
\end{table*}

While the main experiments use the material-fixed subset of the SfC benchmark, the SfC dataset~\cite{TKanekoCVPR2025} also provides material variations under controlled geometry.
Here, we used these variations to evaluate whether reconstruction remains stable under changes in material properties and models.
This analysis differs from the unknown-parameter setting in Appendix~\ref{subsec:relaxing_assumptions}: the physical parameters are provided for each sequence rather than estimated, while either the parameter values or the material model differs from the default elastic setting.
Thus, the goal is to evaluate robustness to material variations rather than physical-parameter estimation accuracy.
We organized the evaluation along two axes: material variation and geometry setting.

\smallskip
\noindent\textbf{Material axis.}
We evaluated three types of material variation from the SfC dataset:
\textit{(i) Young's modulus variation}, where $E$ is varied over
$\{2.5\times10^5, 5.0\times10^5, 1.0\times10^6, 2.0\times10^6, 4.0\times10^6\}$
with the other simulation settings fixed;
\textit{(ii) Poisson's ratio variation}, where $\nu$ is varied over
$\{0.2, 0.25, 0.3, 0.35, 0.4\}$
with the other simulation settings fixed; and
\textit{(iii) material-model variation}, where we evaluated seven material settings: two Newtonian fluids, two non-Newtonian fluids, two plasticine materials, and one sand model, with parameters following the PAC-NeRF dataset~\cite{XLiICLR2023}.

\smallskip
\noindent\textbf{Geometry axis.}
For each material variation, we evaluated reconstruction performance under the following three geometry settings:
\textit{(a) external-shape variation}, where the five external shapes from the SfC dataset are used while the cavity is fixed at the center with size $\left(\frac{2}{3}\right)^3$;
\textit{(b) cavity-size variation}, where the external shape is fixed to a sphere and the cavity size follows the size-variation setting of the main SfC benchmark; and
\textit{(c) cavity-position variation}, where the external shape is fixed to a sphere and the cavity position follows the position-variation setting of the main SfC benchmark.

\subsubsection{Robustness to Young's Modulus}
\label{subsubsec:robustness_young}

Table~\ref{tab:young_material_variation} summarizes reconstruction results under Young's modulus variations across the three geometry settings.
Our method maintains stable performance across stiffness values and consistently improves over SfC-NeRF, indicating robustness to changes in material stiffness.

\subsubsection{Robustness to Poisson's Ratio}
\label{subsubsec:robustness_poisson}

Table~\ref{tab:poisson_material_variation} summarizes reconstruction results under Poisson's ratio variations across the three geometry settings.
Our method maintains stable performance across the evaluated values and consistently improves over SfC-NeRF, indicating robustness to changes in Poisson's ratio.

\subsubsection{Robustness to Material Models}
\label{subsubsec:robustness_material}

Table~\ref{tab:material_model_variation} summarizes reconstruction results under different material models across the three geometry settings.
Our method consistently improves over SfC-NeRF across material categories and geometric variations, indicating that the framework remains effective beyond the elastic setting.

\smallskip
\noindent\textbf{Summary.}
Our method consistently improves over SfC-NeRF across a wide range of elastic parameters and material models.
These results indicate that the proposed hypothesis-space reasoning framework remains robust to material variations beyond the default elastic setting.

\begin{table*}[htbp]
  \centering
  \setlength{\tabcolsep}{4.5pt}
  \small
  \begin{tabular}{lccc cccc}
    \toprule
    & Global fwd. & Local grad. & Total
    & \multicolumn{2}{c}{Size}
    & \multicolumn{2}{c}{Position} \\
    \cmidrule(lr){5-6}
    \cmidrule(lr){7-8}
    Method
    & evals & iters & budget
    & CD$\downarrow$ & F$\uparrow$
    & CD$\downarrow$ & F$\uparrow$ \\
    \midrule
    SfC-NeRF (default)
    & 0 & 1,000 & 1,000
    & 0.165 & 0.135
    & 0.281 & 0.128 \\
    SfC-NeRF (extended)
    & 0 & 2,300 & 2,300
    & 0.167 & 0.136
    & 0.288 & 0.128 \\
    Ours
    & 250 & 2,000 & 2,250
    & \textbf{0.063} & \textbf{0.151}
    & \textbf{0.109} & \textbf{0.144} \\
    \bottomrule
  \end{tabular}
  \vspace{-2mm}
  \caption{Computational budget and reconstruction accuracy on the SfC benchmark.
    \emph{Global fwd. evals} denotes the number of forward-only MPM simulations used by the global stage.
    \emph{Local grad. iters} denotes the number of gradient-based differentiable-simulation iterations.
    For SfC-NeRF, this corresponds to density-optimization iterations; for ours, to Lagrangian particle refinement iterations.
    \emph{Total budget} is reported as the sum of these counts for comparison, although forward-only global evaluations are cheaper than forward--backward gradient iterations and use fewer frames.
    CD values are scaled by $10^3$.}
  \label{tab:runtime}
\end{table*}

\subsection{Runtime Analysis}
\label{subsec:runtime}

We provide additional details for the complexity discussion in Section~\ref{subsec:implementation} of the main paper.
The proposed framework has two computationally different components: global hypothesis optimization uses forward-only MPM simulation~\cite{DSulskyCMAME1994,YHuTOG2018}, whereas Lagrangian particle refinement uses forward--backward differentiable MPM simulation~\cite{YHuICLR2020}.
A single global evaluation requires less computation than a single local refinement iteration because it involves no backward pass and uses only collision-adjacent frames, whereas each local refinement iteration backpropagates through the full observed sequence.
We therefore report the numbers of forward-only global evaluations and gradient-based optimization iterations separately.

SfC-NeRF optimizes densities through differentiable simulation for 1,000 iterations under its default setting.
Our method uses 250 forward-only simulations for coarse-to-fine global hypothesis optimization, followed by 2,000 iterations of Lagrangian particle refinement.
To assess whether the improvement can be explained simply by a larger optimization budget, we also evaluated an extended SfC-NeRF baseline with 2,300 optimization iterations, giving a comparable count-based budget to our method.

Table~\ref{tab:runtime} shows that increasing the number of SfC-NeRF iterations does not improve reconstruction accuracy.
In contrast, our method achieves substantially lower CD and higher F-score with a comparable budget.
This indicates that the improvement comes from changing the optimization structure---evaluating explicit structural hypotheses before local refinement---rather than from increasing the number of gradient-based iterations.

We also measured representative wall-clock runtime on a sphere sequence with cavity size $\left(\frac{2}{3}\right)^3$ using a single H100 GPU.
A forward-only global evaluation on collision-adjacent frames takes about 12 s, whereas one forward--backward local refinement iteration over the full observed sequence takes about 48 s on this sequence.
As a result, the 250-evaluation global search accounts for roughly 3\% of the local-refinement runtime.
This supports the practical feasibility of using a moderate global search budget before local refinement.

\section{Qualitative Results}
\label{sec:qualitative}

We provide additional qualitative comparisons that complement Fig.~\ref{fig:qualitative} of the main paper.
These examples show how the proposed framework behaves under cavity position and size variations, as well as interior-topology stress tests.
Supplementary videos for Fig.~\ref{fig:qualitative} and the examples below are available on the \href{https://www.kecl.ntt.co.jp/people/kaneko.takuhiro/projects/sfc-hypothesis/}{project page}.

\subsection{SfC Benchmark}
\label{subsec:qualitative_sfc}

We first present qualitative reconstruction results on the SfC benchmark, which contains controlled variations in cavity position and size.

\subsubsection{Cavity Position Variations}
\label{subsubsec:qualitative_sfc_position}

\begin{figure*}[htbp]
  \centering
  \begin{subfigure}{\linewidth}
    \centering
    \includegraphics[width=\linewidth]{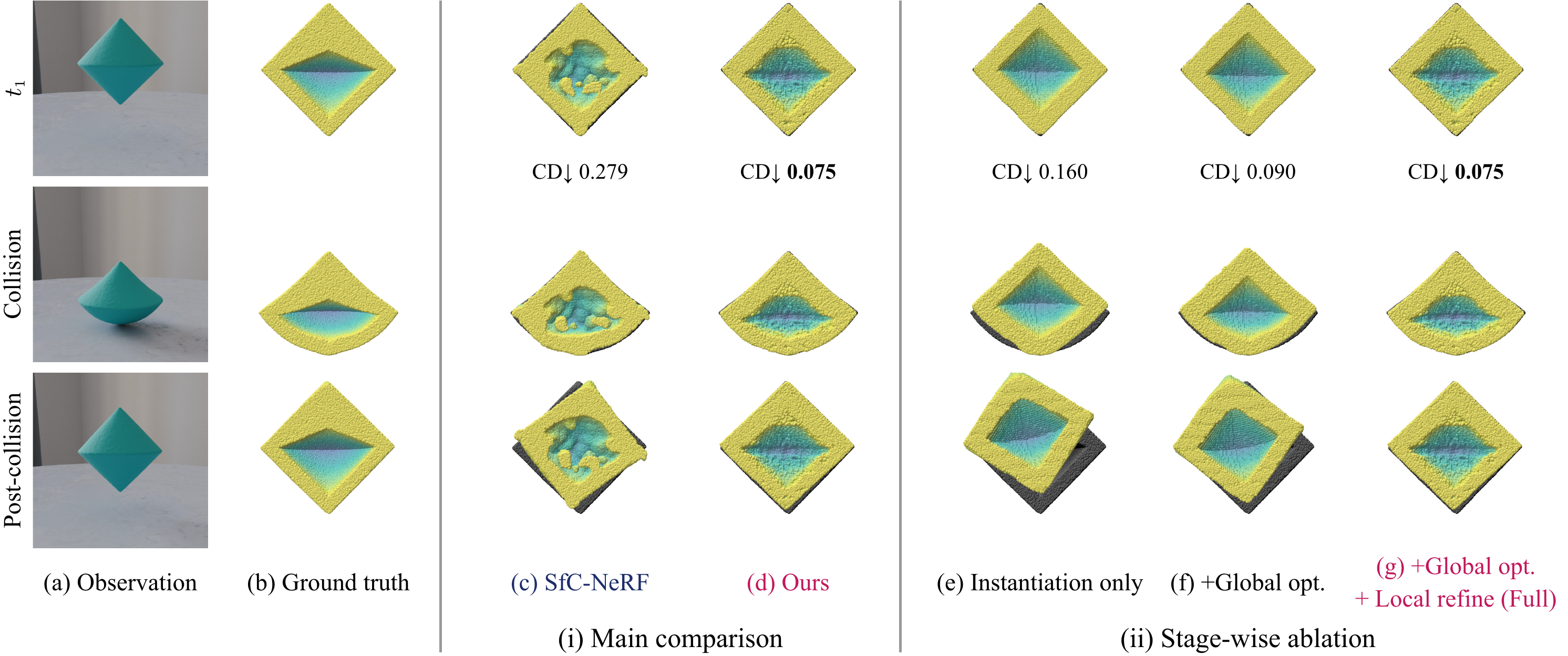}
    \caption{Bicone}
  \end{subfigure}

  \smallskip

  \begin{subfigure}{\linewidth}
    \centering
    \includegraphics[width=\linewidth]{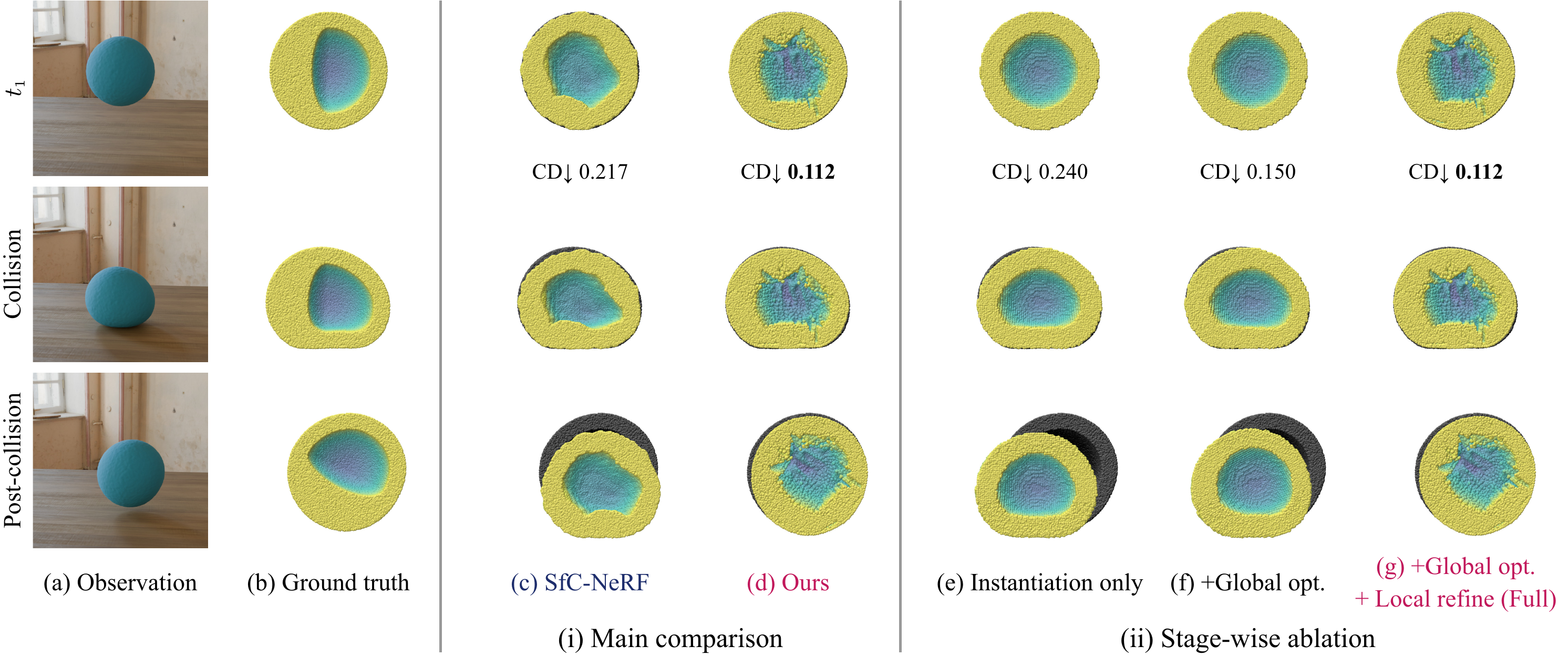}
    \caption{Sphere}
  \end{subfigure}

  \caption{Qualitative reconstruction results under cavity position variation on the SfC benchmark.
    We show examples for \textit{bicone} and \textit{sphere}.
    For each object, we show observed images, ground-truth cross-sections, and reconstructions from different methods and stages.
    Block (i) shows the main comparison between SfC-NeRF and our method, while block (ii) visualizes the stage-wise effects of cavity instantiation, global hypothesis optimization, and Lagrangian particle refinement.
    Rows correspond to the initial frame ($t_1$), the collision moment, and a post-collision frame.
    CD is reported for the initial reconstruction.
    Ground-truth cross-sections are shown in gray.
    Corresponding supplementary videos are available on the \href{https://www.kecl.ntt.co.jp/people/kaneko.takuhiro/projects/sfc-hypothesis/}{project page}.}
  \label{fig:qualitative_position}
\end{figure*}

Figure~\ref{fig:qualitative_position} shows representative results for two objects under cavity position variation.
SfC-NeRF often reconstructs fragmented or irregular cavity structures, which can lead to post-collision dynamics that deviate from the observation.
In contrast, our method reconstructs more coherent internal structures that better match the ground truth.
The stage-wise results further show how each component contributes: cavity instantiation provides an initial structure, global hypothesis optimization improves the coarse cavity configuration, and Lagrangian particle refinement adjusts local geometry.
These differences become more apparent in post-collision frames, where inaccurate cavity configurations lead to different deformation patterns.

\subsubsection{Cavity Size Variations}
\label{subsubsec:qualitative_sfc_size}

\begin{figure*}[htbp]
  \centering
  \begin{subfigure}{\linewidth}
    \centering
    \includegraphics[width=\linewidth]{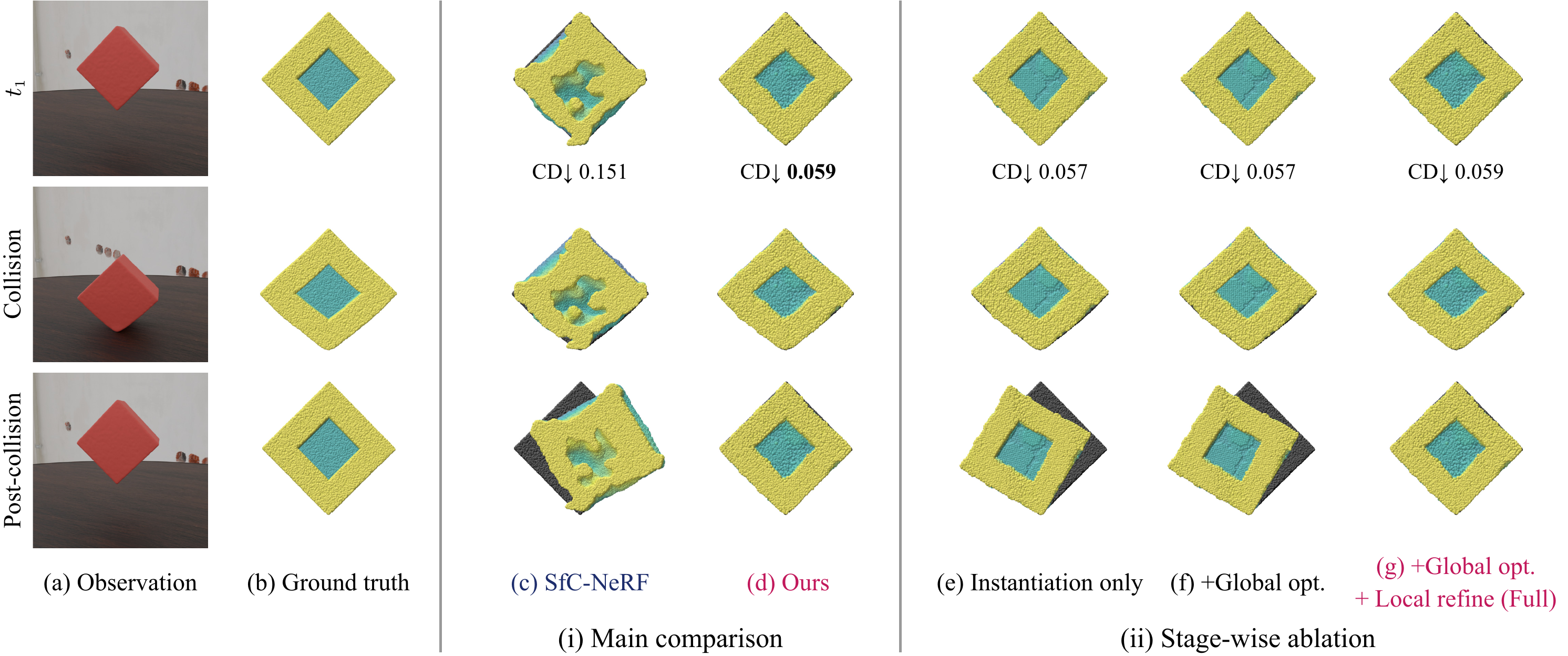}
    \caption{Cube}
  \end{subfigure}

  \smallskip

  \begin{subfigure}{\linewidth}
    \centering
    \includegraphics[width=\linewidth]{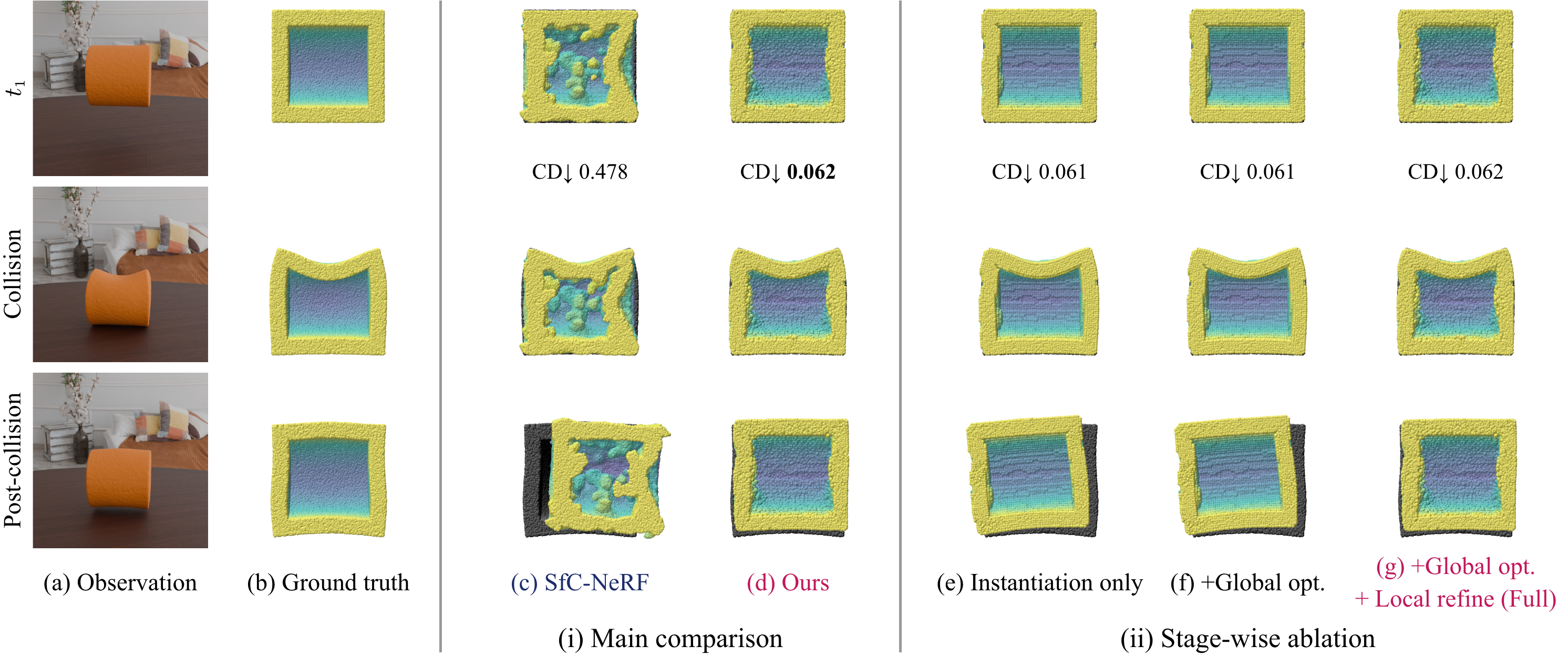}
    \caption{Cylinder}
  \end{subfigure}

  \caption{Qualitative reconstruction results under cavity size variation on the SfC benchmark.
    We show examples for \textit{cube} and \textit{cylinder}.
    For each object, we show observed images, ground-truth cross-sections, and reconstructions from different methods and stages.
    Block (i) shows the main comparison between SfC-NeRF and our method, while block (ii) visualizes the stage-wise effects of cavity instantiation, global hypothesis optimization, and Lagrangian particle refinement.
    Rows correspond to the initial frame ($t_1$), the collision moment, and a post-collision frame.
    The coarse cavity geometry is largely captured after instantiation, while Lagrangian particle refinement improves local geometry and post-collision consistency.
    CD is reported for the initial reconstruction.
    Ground-truth cross-sections are shown in gray.
    Corresponding supplementary videos are available on the \href{https://www.kecl.ntt.co.jp/people/kaneko.takuhiro/projects/sfc-hypothesis/}{project page}.}
  \label{fig:qualitative_size}
\end{figure*}

Figure~\ref{fig:qualitative_size} shows representative results for two objects under cavity size variation.
In this setting, cavity instantiation already captures the coarse cavity extent, leading to smaller differences across stages than in the position-variation setting.
Nevertheless, Lagrangian particle refinement improves local geometric consistency and leads to more stable post-collision dynamics.
These results suggest that, under cavity-size variation, refinement mainly improves physical consistency and local geometry rather than resolving a large positional ambiguity.

\subsection{Interior-Topology Stress Tests}
\label{subsec:qualitative_topology}

\begin{figure*}[htbp]
  \centering
  \includegraphics[width=\linewidth]{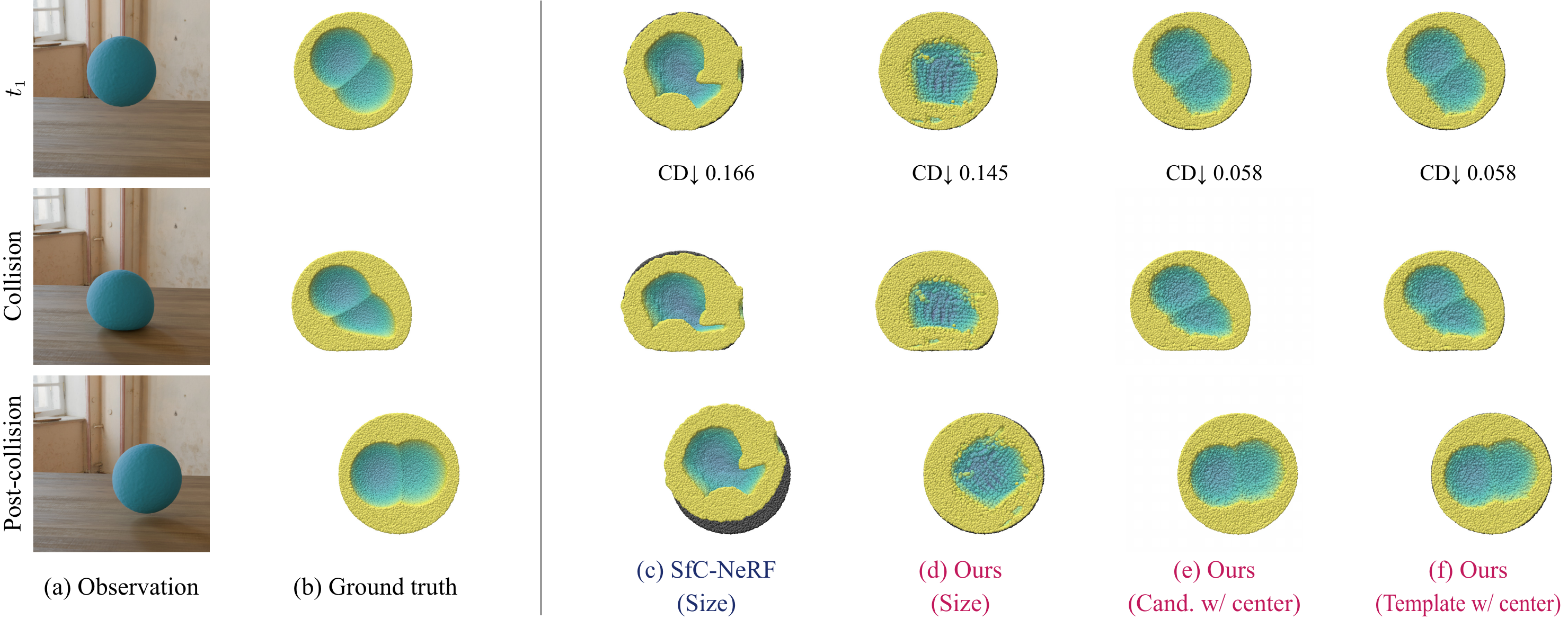}
  \caption{Qualitative results for the symmetric subset of the interior-topology stress tests.
    We show a representative \textit{snowman} cavity.
    Columns correspond to (a) observation, (b) ground truth, (c) SfC-NeRF under the known-size setting, (d) our method under the known-size setting, (e) our method with a set of candidate templates and a known center, and (f) our method with a known template and center.
    Rows correspond to the initial frame ($t_1$), the collision moment, and a post-collision frame.
    The visualization follows the hypothesis-space settings in Table~\ref{tab:topology} of the main paper.
    Stronger structural assumptions constrain the hypothesis space and improve reconstruction consistency.
    Corresponding supplementary videos are available on the \href{https://www.kecl.ntt.co.jp/people/kaneko.takuhiro/projects/sfc-hypothesis/}{project page}.}
  \label{fig:qualitative_topology_symmetric}
\end{figure*}

\begin{figure*}[htbp]
  \centering
  \includegraphics[width=\linewidth]{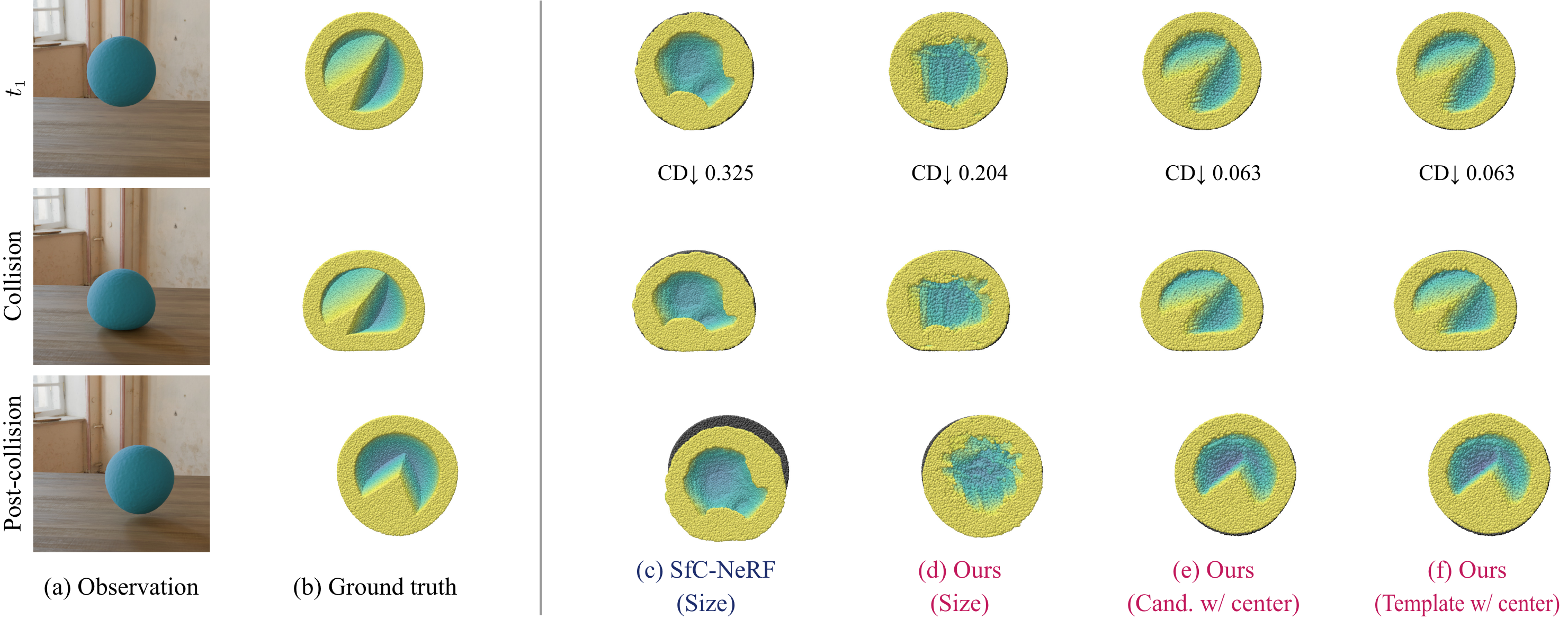}
  \caption{Qualitative results for the asymmetric subset of the interior-topology stress tests.
    We show a representative \textit{pacman} cavity.
    Columns correspond to (a) observation, (b) ground truth, (c) SfC-NeRF under the known-size setting, (d) our method under the known-size setting, (e) our method with a set of candidate templates and a known center, and (f) our method with a known template and center.
    Rows correspond to the initial frame ($t_1$), the collision moment, and a post-collision frame.
    The visualization follows the hypothesis-space settings in Table~\ref{tab:topology} of the main paper.
    The asymmetric cavity introduces a preferred orientation, making this example useful for visualizing how candidate templates and orientation refinement affect reconstruction.
    Corresponding supplementary videos are available on the \href{https://www.kecl.ntt.co.jp/people/kaneko.takuhiro/projects/sfc-hypothesis/}{project page}.}
  \label{fig:qualitative_topology_asymmetric}
\end{figure*}

\begin{figure*}[htbp]
  \centering
  \includegraphics[width=\linewidth]{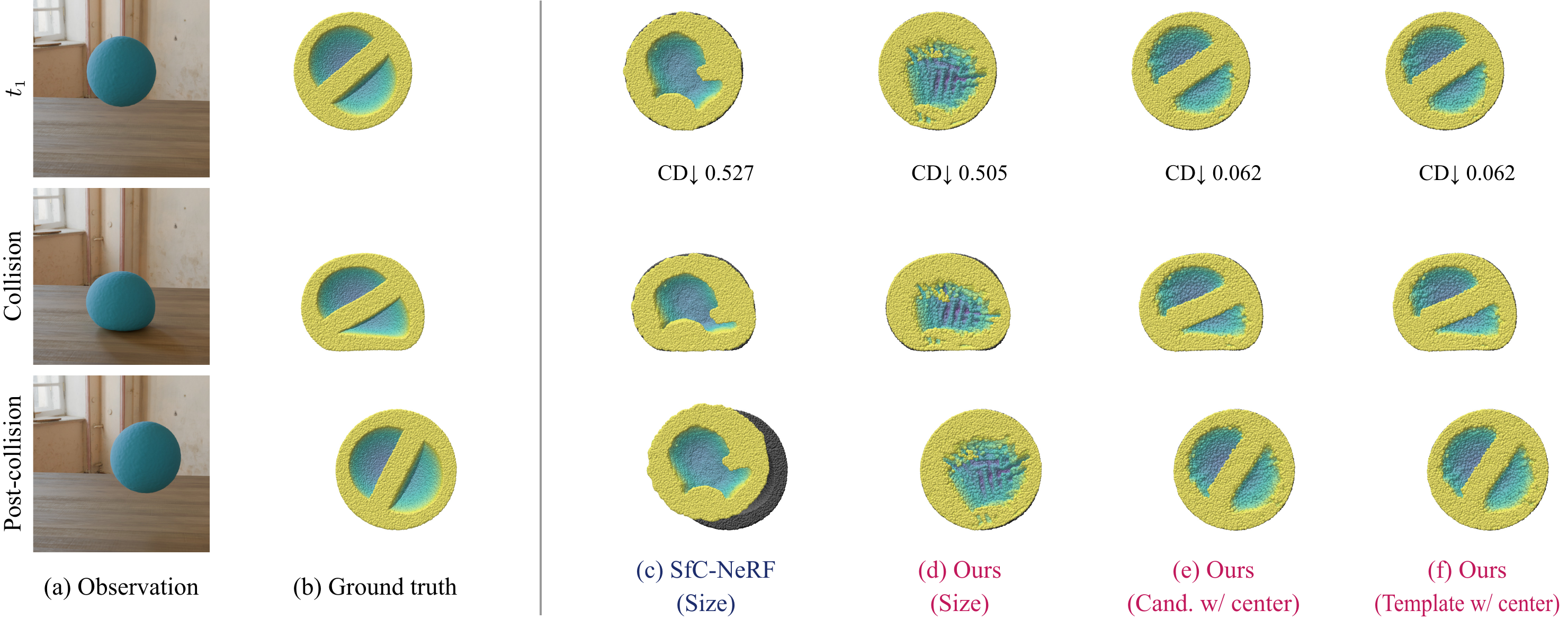}
  \caption{Qualitative results for the complex subset of the interior-topology stress tests.
    We show a representative \textit{split sphere} cavity.
    Columns correspond to (a) observation, (b) ground truth, (c) SfC-NeRF under the known-size setting, (d) our method under the known-size setting, (e) our method with a set of candidate templates and a known center, and (f) our method with a known template and center.
    Rows correspond to the initial frame ($t_1$), the collision moment, and a post-collision frame.
    The visualization follows the hypothesis-space settings in Table~\ref{tab:topology} of the main paper.
    Exact reconstruction remains challenging for complex cavities from collision observations alone, but hypothesis-space reasoning produces more coherent structures and more physically plausible deformation dynamics.
    Corresponding supplementary videos are available on the \href{https://www.kecl.ntt.co.jp/people/kaneko.takuhiro/projects/sfc-hypothesis/}{project page}.}
  \label{fig:qualitative_topology_complex}
\end{figure*}

Figures~\ref{fig:qualitative_topology_symmetric}--\ref{fig:qualitative_topology_complex} show representative qualitative results for the interior-topology stress tests, covering symmetric (\textit{snowman}), asymmetric (\textit{pacman}), and complex (\textit{split sphere}) cavities.
Following Table~\ref{tab:topology} of the main paper, we visualized the known-size setting and two stronger structural-assumption settings.
Under the known-size setting, our method produces more coherent cavity structures than SfC-NeRF, although exact reconstruction remains challenging for complex cavities.
With a set of candidate templates and a known center, the global stage can select among structural alternatives before refinement.
Because the candidate set includes the ground-truth template, this setting can produce results close to the known-template setting, suggesting that hypothesis-level selection can identify a plausible structural alternative from the candidate set.
With a known template and center, the remaining ambiguity is mainly orientation, further constraining the reconstruction problem.
These results are consistent with the quantitative trends in Table~\ref{tab:topology} of the main paper: hypothesis-space reasoning improves reconstruction in the interior-topology stress tests, while exact recovery of complex interior topology remains difficult from collision observations alone.

\smallskip
\noindent\textbf{Summary.}
These qualitative examples support the quantitative findings: global hypothesis optimization improves coarse cavity selection, and Lagrangian particle refinement improves local geometry and post-collision consistency.
They also clarify the scope of the framework: stronger structural assumptions can improve reconstruction in the interior-topology stress tests, but complex cavities remain challenging when the observations do not provide sufficient evidence.

\clearpage

\end{document}